\documentclass[10pt,twocolumn,letterpaper]{article}

\usepackage[pagenumbers]{cvpr} 

\usepackage[dvipsnames]{xcolor}
\usepackage{booktabs}

\usepackage{pifont} % for cmark / xmark
\newcommand{\cmark}{\ding{51}}
\newcommand{\xmark}{\ding{55}}

\usepackage{colortbl}
\usepackage{multirow}
\usepackage{multicol}
\usepackage{listings} % Add code blocks
\usepackage{titletoc} % for table of contents of appendix only, should be BEFORE HYPERREF
\usepackage{fancyvrb}
\usepackage{tcolorbox} % Prompts in appendix
\usepackage[accsupp]{axessibility}  % Improves PDF readability for those with disabilities.

\definecolor{lightgray}{rgb}{0.83, 0.83, 0.83}
\definecolor{Gray}{gray}{0.6}
\definecolor{aliceblue}{rgb}{0.94, 0.97, 1.0}
\definecolor{mistyrose}{rgb}{1.0, 0.89, 0.88}
\definecolor{backcolour}{rgb}{0.95,0.95,0.92}

\newcommand{\newpara}[1]{\vspace{2pt}\noindent\textbf{#1}}

\newcommand{\appendixref}[2]{%
  \if\sepappendix1%
    #1% If separate appendix
  \else%
    #2% If together
  \fi%
}

\definecolor{cvprblue}{rgb}{0.21,0.49,0.74}
\usepackage[pagebackref,breaklinks,colorlinks,allcolors=cvprblue]{hyperref}

\def\sepappendix{0}

\title{Lost in Translation, Found in Embeddings:\\Sign Language Translation and Alignment}

\author{
    \setlength{\tabcolsep}{35pt}
    \begin{tabular}{ccc}
    Youngjoon Jang$^{1,2}$ & Liliane Momeni$^{1}$ & Zifan Jiang$^{1,3}$ \\
    Joon Son Chung$^{2}$ & G\"{u}l Varol$^{1,4}$ & Andrew Zisserman$^{1}$
    \end{tabular}
    \vspace{1mm} \\
    {\normalsize $^1$VGG, University of Oxford \quad $^2$KAIST \quad $^3$University of Zurich \quad $^4$LIGM, École des Ponts, IP Paris, UGE, CNRS} \\
    {\small \url{https://www.robots.ox.ac.uk/~vgg/research/litfie/}}
}

\begin{document}
\maketitle
\begin{abstract}
Our aim is to develop a unified model for sign language understanding, that performs sign language translation (SLT) and sign–subtitle alignment (SSA). 
Together, these two tasks enable the conversion of continuous signing videos into spoken language text and also the temporal alignment of signing with subtitles -- both essential for practical communication, large-scale corpus construction, and educational applications.
To achieve this, our approach is built upon three components: (i)~a lightweight visual backbone that captures manual and non-manual cues from human keypoints and lip-region images while preserving signer privacy; (ii)~a Sliding Perceiver mapping network that aggregates consecutive visual features into word-level embeddings to bridge the vision–text gap; and (iii)~a multi-task scalable training strategy that jointly optimises SLT and SSA, reinforcing both linguistic and temporal alignment. 
To promote cross-linguistic generalisation, we pretrain our model on large-scale sign–text corpora covering British Sign Language (BSL) and American Sign Language (ASL) from the BOBSL and YouTube-SL-25 datasets. With this multilingual pretraining and strong model design,
we achieve state-of-the-art results on the challenging BOBSL (BSL) dataset for both SLT and SSA. 
Our model also demonstrates robust zero-shot generalisation and finetuned SLT performance on How2Sign (ASL), highlighting the potential of scalable translation across different sign languages.
\end{abstract}
\section{Introduction}
\label{sec:intro}
Sign languages are fully-fledged natural languages with their own grammatical structures.
They convey meaning not only through manual gestures such as handshapes, movements and positions, but also through non-manual cues including facial expressions, lip movements and body posture. Given their multi-modal and spatio-temporal nature, understanding sign languages requires interpretation of complex visual and contextual information.

The focus of this work is on developing a unified model that performs sign language translation (SLT) and sign-subtitle alignment (SSA), two fundamental tasks for sign language understanding. SLT aims to \emph{translate} sign language videos into spoken language text, facilitating communication between deaf communities with non-signers. 
SSA, in turn, \emph{aligns} continuous signing videos with spoken language text. Since the temporal correspondence between signing and subtitles is often noisy by nature~\cite{muller-etal-2023-findings}, accurate alignment is challenging but crucial for constructing large-scale parallel corpora and downstream applications such as sign language education and sign segment retrieval.

\begin{figure}
    \centering
    \includegraphics[width=1.0\linewidth]{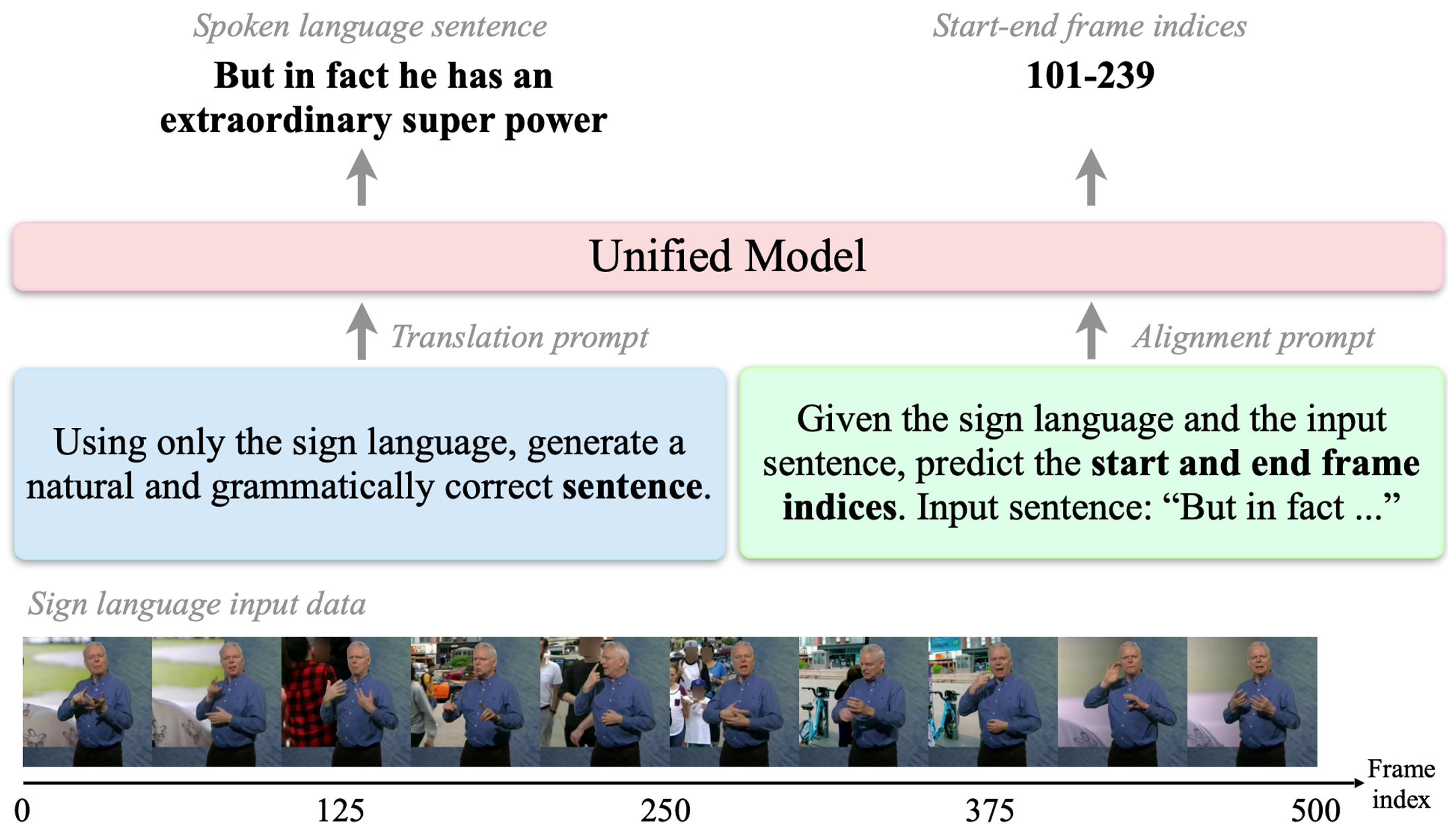}
    \vspace{-7mm}
    \caption{
        \textbf{A unified sign language understanding model.}
        Given signing data, our model performs both SLT and SSA, guided by textual prompts. For both tasks, a 500-frame (20s at 25 fps) video is used as input.
        In SLT mode, the model receives the sign video with frame-level timestamps specifying the region of interest (not shown for clarity), and generates a spoken language translation for that segment.
        In SSA mode, the model takes the sign video, a target sentence along with its audio-aligned timestamps (if available), and predicts the timestamps where the sentence is signed. 
    }
    \vspace{-3mm}
    \label{fig:teaser}
\end{figure}

Recent methods for SLT build on the strengths of large language models (LLMs) pretrained for text translation~\cite{wong2024sign2gpt,jang2025lost,li2025uni,zhou2023gloss}.
A common framework is to proceed in two stages: first, train a visual backbone on datasets with glosses or pseudo-glosses\footnote{We borrow the linguistic \textit{gloss} term and refer to sign-level translations in
free-form English as glosses.}, this backbone maps from video frames to an embedding vector;
and second, train a mapping network from this embedding to the input language space of the LLM, whilst finetuning the LLM for the SLT task.

While this framework has allowed great progress, its performance
depends on (i)~the scale of the training data,
(ii)~the generality of the training data (e.g.~the number and variety of signers), and (iii)~end-to-end training. 
In the first stage of visual backbone training, annotations are often only available at small scale or for a few signers, limiting model generality.
For instance, sign language datasets Phoenix14T~\cite{camgoz-slt}, How2Sign~\cite{duarte2021how2sign} and BOBSL~\cite{Albanie2021bobsl} 
contain only 9, 11, and 39 signers, respectively.
In the second stage, the visual backbone is often frozen~\cite{jang2025lost}, and the mapping network is usually a simple MLP, which leads to a mis-match with the language space of the LLM.
Similarly, SSA suffers from limited-scale annotation, as manual alignment is extremely time-consuming -- an expert requires around 10–15 hours to align subtitles for a one-hour video~\cite{Bull21}.

In this work, we make a number of contributions that 
enable both end-to-end training and better generality in training. 
First, we introduce a
lightweight visual backbone that ingests human body pose keypoints and lip
features. This has two significant advantages: first, privacy and domain generality -- the keypoints remove
visual `nuissance factors' such as clothing and lighting, and the lip features capture non-manual aspects of the signs; and second, 
its lightweight design allows for end-to-end finetuning -- improving performance and facilitating
domain adaptation for downstream tasks. The backbone is pretrained on isolated sign language recognition with pseudo-gloss,
and then further finetuned on web-scale data to achieve generality over signers.

Second, we introduce the `Sliding Perceiver' mapping network, a Perceiver-based~\cite{alayrac2022flamingo,jaegle2021perceiver} network designed to capture local compositionality in sign language. Conventional frame-wise mappings, such as MLPs, fail to model short sequences of gestures that correspond to single words. The Sliding Perceiver addresses this by summarising features within a local sliding window into a single latent, which moves across the sequence to produce word-level embeddings. This compact representation enables better alignment between visual features and linguistic units.

Third, we integrate these components with an LLM into a single model for SLT and SSA, where the target task is 
specified by a prompt (see \cref{fig:teaser}). This joint training enhances inference capabilities.~Through this design, our model achieves state-of-the-art performance on SLT for BOBSL~\cite{albanie2021bbc} and How2Sign~\cite{duarte2021how2sign}, while simultaneously obtaining state-of-the-art in SSA on BOBSL -- highlighting our model generalises across multiple tasks and languages (British and American Sign Languages). These results demonstrate the effectiveness of combining a lightweight backbone, a sign-to-language mapping network, and LLM-based multitask learning. 

\section{Related Works}
\label{sec:related_works}
\newpara{Isolated sign language recognition (ISLR)} classifies pre-segmented video clips into gloss labels -- a task that has significantly advanced over the past decade with the development of deep spatiotemporal architectures~\cite{huang2015sign,camgoz2016using,carreira2017quo,liu2022video,liu2020disentangling,yan2018spatial} and large-scale datasets~\cite{camgoz2016bosphorussign,vaezijoze2019ms,Albanie20,li2020word,sincan2020autsl}. Methods broadly fall into two main categories: (i) appearance-based methods, that directly process RGB frames, and (ii) pose-based approaches. While appearance models like I3D~\cite{carreira2017quo} and the Video Swin Transformer~\cite{liu2022video} have pushed the state-of-the-art~\cite{vaezijoze2019ms,Albanie20,li2020word,prajwal2022weakly,raude2024,jang2025lost}, pose-based approaches~\cite{laines2023isolated,Tunga_2021_WACV,Bohacek_2022_WACV,Hosain_2021_WACV} offer efficiency, signer anonymity, and robustness to background, lighting, and signer variations. However, pose-based models often lag in accuracy, as keypoints alone can omit rich visual details. In this paper, we propose a lightweight ISLR backbone that combines holistic body keypoints from MediaPipe~\cite{lugaresi2019mediapipe} with a grayscale lip-region stream, effectively modeling both manual and non-manual articulations. 
This design achieves a strong balance between accuracy and efficiency, outperforming the Video Swin Transformer~\cite{prajwal2022weakly}, and serves as an effective foundation for downstream sign language translation and alignment tasks.

\begin{figure*}[t]
  \centering
  \includegraphics[width=\linewidth]{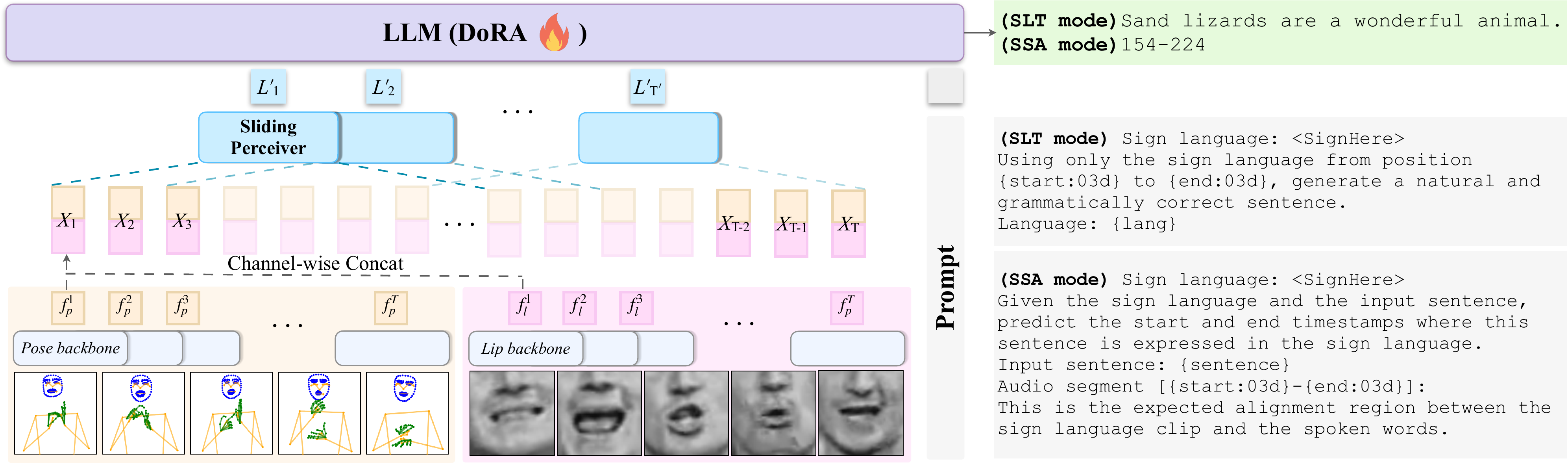}
  \vspace{-7mm}
  \caption{\textbf{Method overview.} Given a 20-second long sign language video, our model extracts two complementary modalities: (i) holistic body keypoints and (ii) grayscale lip-region sequences.
  Each modality is processed by its respective backbone using a sliding-window (window size = $24$, stride = $2$) approach to produce temporally dense pose and lip features, denoted as $f_p$ and $f_l$.
  These features are concatenated along the channel dimension and passed into the Sliding Perceiver, which aggregates local visual information into a compact latent sequence $L'$.
  The latent representation  $L'$ is then inserted within a task-specific prompt (at \texttt{<SignHere>} token) and fed into a pretrained LLM to perform one of two downstream tasks. 
  In SLT mode, the model generates spoken-language text corresponding to the signing video in the region of interest (denoted by frame indices at the \texttt{\{start\}} and \texttt{\{end\}} tokens in the prompt).
  In SSA mode, the model predicts the start and end timestamps, as frame indices, where the provided sentence (inserted at the \texttt{\{sentence\}} token in the prompt) is being signed. Note that the \texttt{\{lang\}} field in the SLT prompt specifies the target language using its ISO 639 code.}
  \label{fig:main} 
  \vspace{-2mm}
\end{figure*}

\newpara{Continuous sign language recognition (CSLR)} transcribes continuous signing videos into gloss sequences. 
These intermediate representations attempt to preserve the unique linguistic and morphological structure of sign language~\cite{sutton1999linguistics} but are limited~\cite{muller-etal-2023-considerations}. 
While CSLR benchmarks like PHOENIX~\cite{Phoenix} and CSL-Daily~\cite{zhou2021improving} provide gloss-annotated data, these annotations are labor-intensive and often lack precise frame-level segmentation. Consequently, most CSLR models employ the Connectionist Temporal Classification (CTC) loss~\cite{ctc} to learn monotonic alignments between visual features and gloss sequences~\cite{cheng2020,Wei2023,CoSign,zuo2022c2slr,Hao2021,Min_2021_ICCV,zheng2023cvt,ahn2024slowfast,jang2023self}.  Recent advances are moving beyond conventional frame-to-gloss modeling. For instance,~\cite{raude2024} explores stronger contextual reasoning by framing CSLR as a video-to-text retrieval problem.

\newpara{Automatic sign language translation (SLT)} aims to translate sign language videos into spoken language sentences. Current SLT research is broadly divided into gloss-based and gloss-free approaches.
Gloss-based SLT uses intermediate gloss predictions from CSLR as explicit input or auxiliary supervision for the translation model~\cite{camgoz-slt,camgoz2020sign,chen2022two,zhou2021improving,chen2022simple,yin-read-2020-better,ye-etal-2023-cross,zhang2023sltunet,stmc-2021}. In contrast, gloss-free SLT learns a direct video-to-sentence mapping~\cite{WMT23SLT_knowcomp,muller-etal-2023-findings}. In this work, we adopt the more scalable, gloss-free approach. This allows us to leverage large-scale video-text pairs without costly gloss annotations, achieving strong, open-vocabulary generalisation across diverse signers and languages.

Recent gloss-free methods vary in visual pretraining and LLM adaptation. Some use features from pretrained ISLR backbones~\cite{Albanie2021bobsl,slt-how2sign-wicv2023} or contrastively align visual and textual features~\cite{lin-etal-2023-gloss,zhou2023gloss,ye2024improving,jiao2024visual}. Others integrate LLMs, either by feeding discrete visual tokens into frozen models~\cite{touvron2023llama}, or by finetuning them with visual representations~\cite{uthus2023youtubeasl,sandoval-castaneda-etal-2023-ttics,rust2024towards,jiao2024visual,jang2025lost}, potentially using parameter-efficient methods like LoRA~\cite{hu2022lora,wong2024sign2gpt}. Additionally, lightweight, pose-based backbones which can be trained end-to-end for translation have shown strong performance~\cite{li2025uni,fish2025geo}. Building on this trend, our work employs a pretrained LLM (Flan-T5~\cite{chung2024scaling}) and finetunes it for sign language translation and alignment, updating the visual backbone in an end-to-end manner.
 
\newpara{Sign–subtitle alignment (SSA)} temporally matches continuous sign language videos with their corresponding subtitle units. Early methods~\cite{farhadi2006aligning} used sparse correspondences between video and subtitles, often assuming that the word order in subtitles directly reflects the signing sequence. Later studies on related tasks like sign segmentation~\cite{farag2019learning,renz2021sign}, diarisation~\cite{gebrekidan2013automatic,gebre2014motion,albanie2021signer}, and active signer detection~\cite{cherniavsky2008activity,borg2019sign,moryossef2020real,shipman2017speed} focus on temporal granularity at the word or active-signer level rather than aligning full subtitle units. While \cite{bull2020automatic} segments videos into sentence-like units, these still require translation to be aligned. More recently, work has tackled large-scale broadcast data~\cite{Bull21} or incorporated linguistic knowledge~\cite{jang2025deep}. To our knowledge, we present the first framework to perform SSA using an LLM, enabling more accurate and scalable alignment.

The most related work~\cite{tanzer2024fleurs} jointly trains SLT and caption alignment, predicting caption timing from video only, but attains just 24.3\% frame accuracy, revealing the limitations of relying solely on visual cues.
In contrast, our model leverages both video and subtitles to localise signing segments, enabling practical generation of sign-aligned subtitles from audio-aligned transcripts.

\section{Method}
\label{sec:method}
Our model is illustrated in \cref{fig:main}. Given a 20-second signing video and a task-specific prompt, the model produces outputs for both SLT and SSA.
In this section, we first describe the architecture (\cref{subsec:architecture}), including the visual backbone and the Sliding Perceiver mapping network, followed by an overview of our training stages (\cref{subsec:training-overview}).
We then dive deeper into the backbone pretraining (\cref{subsec:backbone}), and downstream SLT and SSA objectives to achieve robust recognition and temporal grounding (\cref{subsec:downstream}). 

\subsection{Architecture}
\label{subsec:architecture}
Here we describe our visual feature processing: the pose backbone, lip backbone, and Sliding Perceiver.

\begin{figure}[t]
  \centering
  \includegraphics[width=\linewidth]{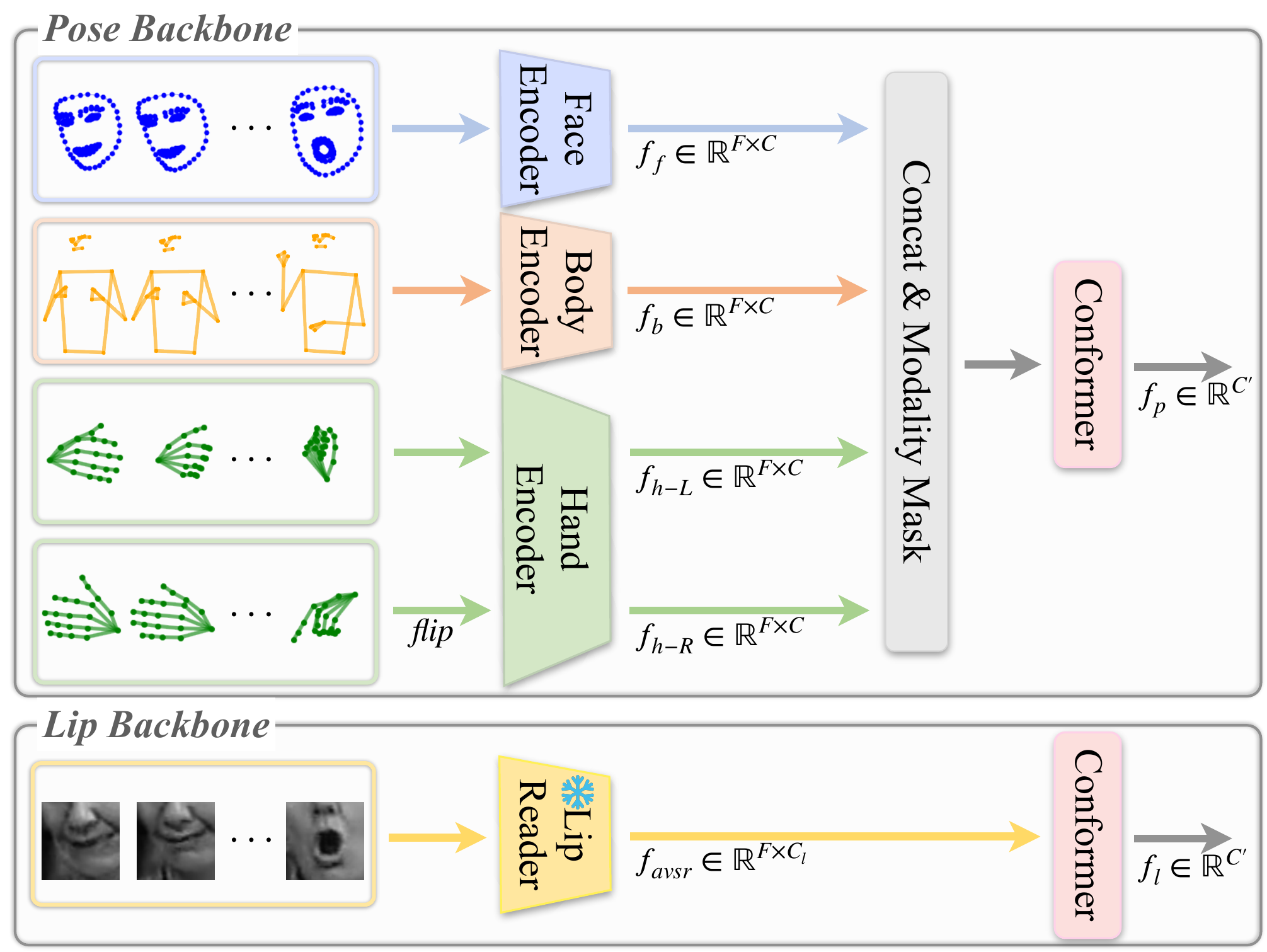}
  \vspace{-7mm}
  \caption{\textbf{Visual backbone.} The model has two streams: pose and lip backbones. The pose backbone processes $F$ frames of 3D keypoints, divided into face, body, and hand articulators, encoding each to produce articulator-specific features ($f_f$, $f_b$, $f_{h\text{-}L}$, $f_{h\text{-}R}$). Up to two of these features are randomly masked during training to improve robustness. The features are then concatenated along the channel dimension, and passed through a Conformer to obtain the pose embedding.
  The lip backbone also processes $F$ grayscale video frames of the signer lip-region via a frozen pretrained lip reader, followed by a Conformer that outputs the lip embedding.
  }
  \label{fig:main_backbone}
  \vspace{-4mm}
\end{figure}

\newpara{Pose backbone.}
As illustrated on the top half of \cref{fig:main_backbone}, we employ the 3D keypoints provided by MediaPipe~\cite{lugaresi2019mediapipe}.
In total, 203 keypoints are used: 128 for the face, 33 for the body, and 21 for each hand, all normalised by setting the shoulder length to 1 and centred at the origin~\cite{jiang-etal-2024-signclip}.
Keypoints are grouped into four articulators—face, body, left hand, and right hand—and processed by articulator-specific encoders, each comprising a linear projection to $C_\text{in}$ channels followed by a 4-layer Adaptive Graph Convolutional Network (AGCN)~\cite{shi2019two,li2018adaptive}, with Squeeze-and-Excitation (SE) blocks~\cite{hu2018squeeze} for channel-wise recalibration.
For efficiency, the left and right hand encoders are shared; to align the graph structures, the right-hand keypoints are flipped along the $x$-axis before being input to the shared encoder.
Through the 4-layer AGCN, the channel dimensions are progressively scaled by factors of $[2, 1, 1.5, 2]$, resulting in a final feature size of $C = 6C_\text{in}$ for each keypoint.
The resulting video features therefore have a shape of $F \times J_{m} \times C$, where $F$ is the number of frames and $J_{m}$ is the number of keypoints for each articulator.

Unlike typical action recognition tasks that mainly rely on a few body joints, sign language involves highly detailed facial, body, and finger articulations. 
Thus, naïvely applying spatio-temporal average pooling, as in action recognition, leads to severe information loss. 
Instead, for each frame, we aggregate information across all joints by reshaping the feature map of size $F \times J_{m} \times C$ to $F \times (J_{m} \times C)$, followed by a learnable linear projection that maps it to $F \times C$. 
This enables the model to integrate detailed joint interactions rather than collapsing them through averaging.
The four articulator-specific features ($F \times C$ each) are concatenated along the channel dimension to form a unified representation of $F \times 4C$. 
The resulting feature is then projected to $C'$ dimensions and fed into a Conformer~\cite{gulati20_interspeech} encoder.
The Conformer consists of 4 blocks followed by an attentive pooling layer. 
The attentive pooling computes a mean attention map across all blocks and multiplies it with the output of the final block. 
Temporal averaging is then applied to obtain the final pose feature $f_{p} \in \mathbb{R}^{C'}$. 

\newpara{Lip backbone.}
The purpose of this encoder is to capture
non-manual articulations, such as mouthings (and the tongue and the teeth), that are not fully represented by keypoints.
As shown on the bottom of \cref{fig:main_backbone}, the lip backbone processes grayscale video clips of the lip-region and extracts frame-wise features $f_{avsr} \in \mathbb{R}^{F \times C_l}$ using a pretrained lip-reading network~\cite{ma2023auto}.
To leverage the generalised visual–phonetic representations learned from large-scale audio-visual data~\cite{afouras2018lrs3}, we keep the lip reader frozen during training and only update the subsequent 4-layer Conformer encoder. The Conformer further refines the temporal dynamics of the lip motion and produces the lip feature $f_{l}$. 

\begin{figure}[t]
  \centering
  \includegraphics[width=\linewidth]{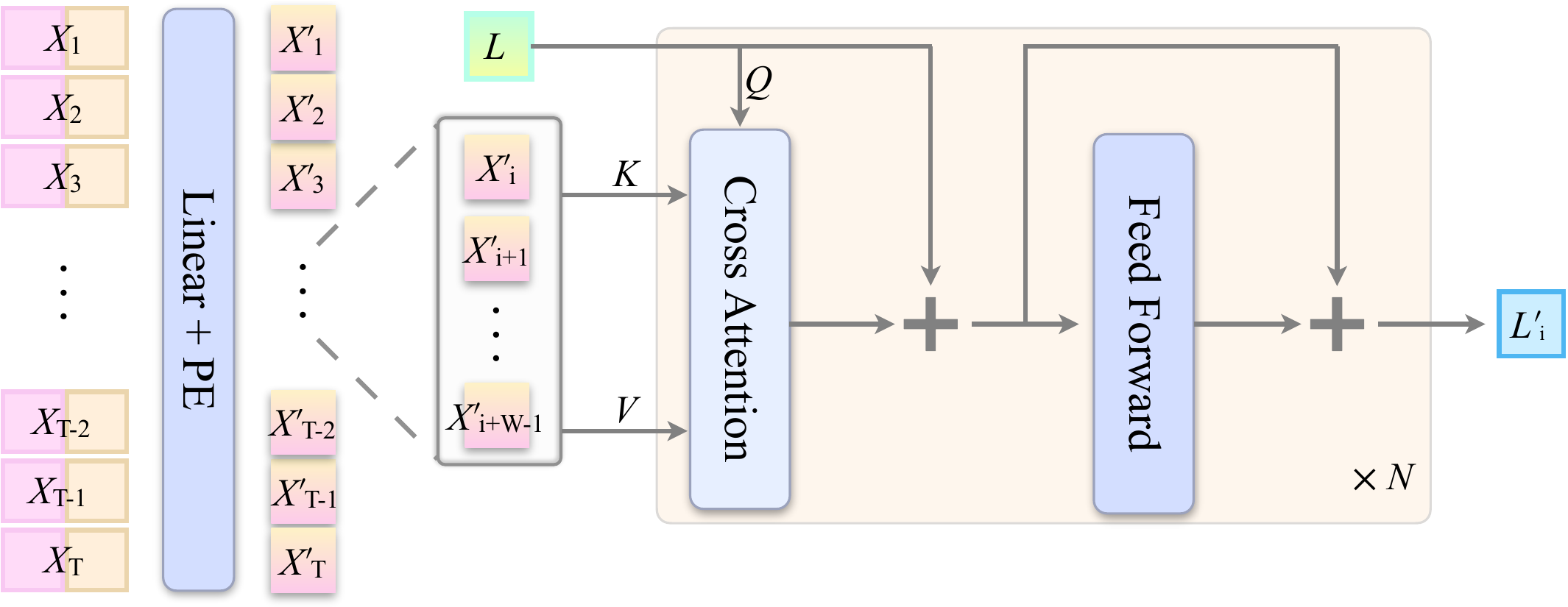}
  \vspace{-7mm}
  \caption{\textbf{Sliding Perceiver.} The input sequence $X$
  is linearly projected and combined with learnable temporal positional encodings. The resulting sequence $X'$ is then processed in a sliding-window manner. Each window is passed through $N$ stacked cross-attention and feed-forward layers with a learnable latent vector $L$.}
  \label{fig:sliding_perceiver}
  \vspace{-4mm}

\end{figure}

\newpara{Sliding Perceiver.}
Unlike the standard Perceiver resampler~\cite{alayrac2022flamingo,jaegle2021perceiver}, which processes the entire sequence globally, our Sliding Perceiver captures local visual context, crucial for sign language where adjacent frames form a single lexical unit. 
As shown in~\cref{fig:sliding_perceiver}, the Sliding Perceiver takes the multimodal feature sequence $X$ as input, where $X$ is obtained by concatenating the pose and lip features ($f_{p}$, $f_{l}$) along the channel dimension.
The sequence is first linearly projected to match the LLM embedding dimension $D$, and a learnable global temporal positional encoding is added to produce $X'$. Then, $X'$ is processed in a sliding-window manner through $N$ stacked cross-attention and feed-forward layers, where a learnable latent vector $L$ serves as the query and each windowed segment of $X'$ provides keys and values. This aggregates local context into latent representations, which are fed into the LLM with the text prompt for downstream tasks.

\subsection{Training overview}
\label{subsec:training-overview}
The pose and lip backbones are first pretrained independently for sign classification (ISLR), enabling each to capture distinct aspects of signing. For the primary tasks of SLT and SSA, these pretrained visual backbones are combined with a Sliding Perceiver and a \textit{Flan-T5-XL} LLM~\cite{chung2024scaling} (see \cref{fig:main}). The main training pipeline next consists of two stages: large-scale pretraining and finetuning. Note that the LLM is efficiently tuned with DoRA~\cite{liu2024dora} adapters.

The large-scale pretraining focuses on English-related sign languages, using 2,700 hours of videos from the ASL and BSL subsets of YouTube-SL-25~\cite{tanzer2024youtube} and the BOBSL dataset~\cite{Albanie2021bobsl} (BSL). 
To avoid overfitting on the easier SSA task, we employ a three-stage curriculum learning strategy.
First, we freeze the visual backbone and train only the mapping network (Perceiver and DoRA) on SLT to align visual features with the LLM. Second, the full model (including the visual backbones, Perceiver and DoRA), is trained exclusively on SLT. Third, the full model is trained on a multitask objective, mixing SLT and SSA samples to encourage shared representations across translation and alignment tasks. Finally, the entire model is finetuned on the specific target dataset, while maintaining multitask training. This allows the model to adapt to domain-specific signing styles and vocabulary, while retaining the strong, general representations acquired from pretraining.

\subsection{Visual backbone ISLR training losses}
\label{subsec:backbone}
\newpara{Classification loss.}
The pose and lip backbones are trained independently.
The output features from the pose backbone ($f_{p}$) and lip backbone ($f_{l}$) are each passed through their respective classifier heads to produce the logits $\hat{y}_{p}$ and $\hat{y}_{l}$, which are then trained against the pseudo-gloss label $y$ using the standard cross-entropy loss.

\newpara{Articulator auxiliary loss (for pose backbone).}
To encourage the different articulators to learn independently while sharing a common representation, we introduce an auxiliary Conformer.
The four feature streams ($f_f$, $f_b$, $f_{h\text{-}R}$, $f_{h\text{-}L}$) are processed independently through the shared Conformer, and the resulting features are fed into the classifier head shared with the pose backbone to produce logits for each articulator (see \cref{fig:app:pose_backbone_full} in the supp.\ mat.). 
These logits are also supervised with the pseudo-gloss labels $y$:

{
\setlength{\abovedisplayskip}{3pt}
\setlength{\belowdisplayskip}{3pt}
\begin{equation*}
\mathcal{L}_{aux} = \frac{1}{N_m} \sum_{m \in \{f, b, h\text{-}R, h\text{-}L\}} \mathcal{L}_{\text{CE}}(\hat{y}_m, y),
\end{equation*}
}
where $N_{m} = 4$ is the total number of articulators, and $\mathcal{L}_{\text{CE}}$ denotes the standard cross-entropy loss.

\newpara{Contrastive loss (for both backbones).}
To enable better alignment of visual features with the pretrained LLM used in downstream tasks, we add a contrastive loss between the visual and textual embedding spaces. Specifically, the pseudo-gloss label $y$ is first converted into a text embedding using the embedding layer of the pretrained LLM, and then projected via a trainable MLP to obtain $f^{\text{proj}}_w$.
Likewise, the visual feature $f_{v}$ (either $f_{p}$ or $f_{l}$) is passed through another MLP projector to produce $f^{\text{proj}}_v$ (find more details in \cref{subsec:app:mlp} of the supp. mat.).
A contrastive loss is then applied between the projected visual and text embeddings:
\begin{equation*}
\setlength{\abovedisplayskip}{3pt}
\setlength{\belowdisplayskip}{3pt}
\mathcal{L}_{contra} = - \frac{1}{B} \sum_{i=1}^{B} 
\log \frac{
\exp(\mathrm{sim}(f^{\text{proj}}_{v_i}, f^{\text{proj}}_{w_i}) / \tau)
}{
\sum_{j=1}^{B} \alpha_{ij} \exp(\mathrm{sim}(f^{\text{proj}}_{v_i}, f^{\text{proj}}_{w_j}) / \tau)
},
\end{equation*}
where $\mathrm{sim}(\cdot,\cdot)$ denotes cosine similarity and $B$ and $\tau$ are batch size and temperature parameters.
The weighting factor $\alpha_{ij}$ handles duplicate class samples within a batch. For samples $i$ and $j$ sharing the same pseudo-gloss label, $\alpha_{ij}$ is the inverse of that label's batch frequency, ensuring each class contributes equally to the loss regardless of its count. Note that the visual feature $f_{v}$ corresponds to $f_{p}$ for the pose backbone, and to $f_{l}$ for the lip backbone.

\newpara{Total training loss.} The overall objective for the pose and lip backbones become:
{
\setlength{\abovedisplayskip}{3pt}
\setlength{\belowdisplayskip}{3pt}
\begin{gather*}
\mathcal{L}_{pose} = \mathcal{L}_{\text{CE}}(\hat{y}_{p}, y) + \mathcal{L}_{aux} + \mathcal{L}_{contra}, \\
\mathcal{L}_{lip} = \mathcal{L}_{\text{CE}}(\hat{y}_{l}, y) + \mathcal{L}_{contra}.
\end{gather*}
}

\subsection{SLT and SSA training losses}
\label{subsec:downstream}
\newpara{Translation loss.}
Conditioned on the SLT prompt and the sign embeddings,  
the LLM generates the spoken language translation, supervised by the standard cross-entropy loss.

\newpara{Alignment loss.}
Given the SSA prompt and the sign embeddings, the model predicts temporal boundaries in a fixed \texttt{start-end} format, where \texttt{start} and \texttt{end} denote the start and end frame indices within the input sequence.
Conventional cross-entropy loss enforces exact token matching but does not reflect how close a predicted timestamp is to the target.
To address this, we use a \emph{soft decoding} mechanism that interprets each predicted digit distribution as an expected numeric value. 
Each digit is restricted to ten tokens (0–9), producing a 10-way softmax $\mathbf{p}_{d} \in \mathbb{R}^{10}$, which is converted to a continuous value $\tilde{y}_{d}$ via soft decoding:
{
\setlength{\abovedisplayskip}{3pt}
\setlength{\belowdisplayskip}{3pt}
\begin{equation*}
\tilde{y}_{d} = \sum_{k=0}^{9} k \cdot \mathbf{p}_{d}(k).
\end{equation*}
}
As each timestamp has six digits (three for \texttt{start}, three for \texttt{end}), the digit-wise L1 loss is computed as follows:
{
\setlength{\abovedisplayskip}{3pt}
\setlength{\belowdisplayskip}{3pt}
\begin{equation*}
\mathcal{L}_{\text{L1}} = \sum_{d=1}^{6} \left| \tilde{y}_{i,d} - y_{i,d} \right|,
\end{equation*}
}
where $y_{i,d}$ is the target digit label.
The final SSA objective combines discrete and continuous components:
{
\setlength{\abovedisplayskip}{3pt}
\setlength{\belowdisplayskip}{3pt}
\begin{equation*}
\mathcal{L}_{\text{SSA}} = \mathcal{L}_{\text{CE}} + \lambda \mathcal{L}_{\text{L1}},
\end{equation*}
}
where $\mathcal{L}_{\text{CE}}$ ensures categorical correctness and $\mathcal{L}_{\text{L1}}$ encourages numerically precise alignment.

\begin{table*}
    \centering
    \setlength{\tabcolsep}{10pt}
    \resizebox{\linewidth}{!}
    {
        \begin{tabular}{l|ccc|ccc|ccc|ccc|c}
        \toprule
        Data & \multicolumn{3}{c|}{Total Dur. (hrs)} & \multicolumn{3}{c|}{Signing Dur. (hrs)} & \multicolumn{3}{c|}{\#Vocab} & \multicolumn{3}{c|}{\#Sentence} & {\#Signers} \\
        \bottomrule
        BOBSL-SENT~\cite{Albanie2021bobsl} & 1,236 & 3 & 31 & 923 & 2.2 & 25.5 & 77,109 & 3,175 & 12,180 & 743,139 & 1,973 & 20,338   & 39 \\
        Youtube-SL-25$^\dagger$~\cite{tanzer2024youtube} & 1,468 & - & - &  833 & - & - & 66,272 & - & - & 441,140 & - & - & 2,583$^*$ \\
        How2Sign~\cite{duarte2021how2sign} & 69.6 & 3.9 & 5.6 &  50.4 & 3.3 & 4.3 & 12,650 & 2.855 & 3,257 & 28,743 & 1,741 & 2,357 & 11 \\
        FLEURS-ASL$^\dagger$~\cite{tanzer2024fleurs} & - & - & 5.2 &  - & - & 5 & - & - & 4,463 & - & - & 1,110 & 5 \\
        \toprule
        \end{tabular}
    }
    \vspace{-0.4cm}
    \caption{\textbf{Dataset statistics.} 
    % Includes total video duration, actual signing duration (i.e., segments where the signer is actively signing), vocabulary size, and number of sentences. 
    % \az{Can we add an estimate of the number of signers in each dataset? Also, where do we give dev vs test sizes?}
    % \jyj{the order of these numbers is train/dev/test. in terms of the number of signer, could we input the number of total signers?}
    Each set of three numbers (except for \#Signer) corresponds to the train, val, and test splits, respectively.
    Vocabulary size is computed after lowercasing, fixing contractions, removing special characters, and lemmatisation. For YouTube-SL-25$^\dagger$, only the ASL and BSL subsets are used. FLEURS-ASL$^\dagger$ refers to the zero-shot devtest split. 
    For YouTube-SL-25$^\dagger$, the number of signers (denoted with $*$) is estimated by the number of unique video channels in the ASL and BSL subsets used in our experiments.
    }
    % For YouTube-SL-25$^\dagger$, the number of signers (denoted with $*$) is estimated by the number of unique video channels 
    % \az{is this the number of channels for the entire dataset, or only for the part we use?}. \jyj{only for the part we use.}}
    \vspace{-0.4cm}
    \label{tab:data_stat}
\end{table*}
\section{Experiments}
\label{sec:experiments}
\subsection{Data and evaluation protocols}
\label{subsec:data}

\cref{tab:data_stat} reports statistics for datasets used in our experiments.

\newpara{BOBSL~\cite{Albanie2021bobsl}} is a large-scale dataset of 1,400+ hours of BSL-interpreted TV footage, with English subtitles corresponding to the audio. 
For the ISLR-training, we follow~\cite{raude2024}, adopting an 8K pseudo-gloss vocabulary with 3.5M training samples from BOBSL-SIGN (each of 24 frames), and corresponding validation and test splits containing 53K and 24K samples, respectively.
During large-scale pretraining, we use automatically generated signing-aligned subtitles~\cite{jang2025deep} instead of audio-aligned ones to provide more accurate supervision, denoted as BOBSL-SENT. 
We filter the dataset to include clips between 1 and 20 seconds, and remove samples with subtitles of more than 80 words.

\newpara{YouTube-SL-25~\cite{tanzer2024youtube}} is a large-scale multilingual sign language dataset containing over 3,000 hours of videos across more than 25 languages, collected from YouTube. 
We use only the ASL and BSL subsets for the large-scale pretraining, and apply the same filtering as for BOBSL.
To reduce noise in the raw subtitles, text cleaning is performed (further details are provided in \cref{subsec:app:text_cleaning} of the supp.\ mat.).

\newpara{How2Sign~\cite{duarte2021how2sign}} is an ASL dataset consisting of approximately 80 hours of instructional video content covering 10 topics. Subtitles are manually aligned to the signing.

\newpara{FLEURS-ASL~\cite{tanzer2024fleurs}} extends the multilingual FLORES~\cite{guzman2019flores,goyal2022flores} and FLEURS~\cite{conneau2023fleurs} benchmarks to ASL, with translations provided by five certified Deaf interpreters. This dataset provides manually aligned subtitles and supports zero-shot, signer-dependent, and signer-independent evaluation for sentence- and discourse-level translation. We process the videos, converting varying frame rates to 25fps.

\newpara{Evaluation metrics.}
For ISLR, we report both per-instance and per-class accuracies, following~\cite{Albanie20}. For SLT, we follow the evaluation protocol of~\cite{jang2025lost}: BLEU-4 (B4)~\cite{bleu}, BLEURT (B-RT)~\cite{sellam2020bleurt}, ROUGE-L (R-L)~\cite{lin-2004-rouge}, CIDEr~\cite{cider}, IoU, and an LLM-based evaluation (LLM). For SSA, we follow~\cite{Bull21}, reporting frame-level accuracy (Acc) and F1-score at IoU thresholds 0.1, 0.25, and 0.50. We further report class-balanced accuracy (CB-Acc) defined as the average per-class recall; its details are provided in \cref{subsec:app:cb_acc} of supp.\ mat. All metrics are higher-is-better.

\subsection{Implementation details}
\label{subsec:implementation}
\newpara{Architecture.}
For the pose backbone, the AGCN has input channels $C_{in}=64$, output channels $C=384$, and a temporal kernel size of 7. The lip backbone uses a pretrained lip-reading model~\cite{ma2023auto} with output dimension 768. Both backbones employ a Conformer with hidden size $C'=512$, 8 attention heads, feed-forward multiplier 4, convolution expansion 2, and kernel size 5. 
The Sliding Perceiver processes the 250-frame input features $X$ using a window size of 8 and stride of 2, producing temporally compressed outputs of length $T' = 122$. It consists of $N = 2$ layers with a latent dimension of $D = 2{,}048$, each employing 8 attention heads and a feed-forward multiplier of 4.
The DoRA~\cite{liu2024dora} adapters are configured with $r=32$ and $\alpha=64$.

\newpara{ISLR-pretraining.} 
Both backbones are trained for 20 epochs using the AdamW~\cite{loshchilov2017decoupled} optimizer with a learning rate of $2 \times 10^{-4}$ and a weight decay of $1 \times 10^{-4}$. 
A one-cycle cosine scheduler is employed to adjust the learning rate. 
We use 4 A5000 GPUs with a batch size of 64 per GPU. 
For the contrastive loss, the temperature parameter $\tau$ is set to 0.1.
During pose backbone training, up to 50\% of the articulator modalities (i.e., at most two) are randomly masked to improve robustness against missing visual cues. 

\newpara{Large-scale pretraining.}
In practice, to reduce memory usage, all 500 frames are fed at once through the visual backbone up to the Conformer, while the Conformer applies sliding windows to leverage prior knowledge learned from 24-frame inputs, lowering required memory from 29.6GB to 5.6GB per batch.
We use the AdamW optimizer with a learning rate of $1 \times 10^{-4}$ and a one-cycle cosine scheduler. For the three-stage pretraining, each stage is trained for 100K iterations with a 2K-iteration warm-up. 
For finetuning, we also use AdamW with a learning rate of $5 \times 10^{-5}$ and a one-cycle cosine scheduler. On BOBSL, we train for 100K iterations with a 2K warm-up, while for How2Sign, we train for 20K iterations with a 500-step warm-up. 
Both pretraining and finetuning are conducted on 4 A6000 GPUs and gradient accumulation of 2 steps is applied to effectively increase the batch size. During multitask training in both pretraining and finetuning, we employ a mixed-batch sampling strategy to balance exposure to SLT (80\%) and SSA (20\%) objectives -- we find this ratio empirically yields strong performance on both tasks, often surpassing single-task baselines. The soft decoding loss is weighted by $\lambda = 0.01$. 
The audio segment in the SSA task prompt is included with 50\% probability only when audio-aligned subtitles are available, i.e., for the BOBSL dataset.

\newpara{Inference.}
During inference, the input sign video is sampled differently depending on the task. For SLT, we centre the 20 second input window around the midpoint of the start and end times of the target sentence. For SSA, we centre the 20-second window around the midpoint of the prior timing from audio-aligned subtitles and perform alignment over the full video following the dynamic time warping (DTW) procedure from~\cite{Bull21,jang2025deep} to remove overlaps. Since the LLM does not provide frame-level confidence, we approximate it with a cosine function that peaks at the predicted timestamp center and decreases to zero at the boundaries (refer to \cref{subsec:app:dtw} of supp.\ mat.). The model then decodes the output auto-regressively, using beam search with a beam size of 5.

\begin{table}
    \setlength{\tabcolsep}{13pt}
    \centering
    \resizebox{\linewidth}{!}
    {
        \begin{tabular}{l|cc|cc}
        \toprule
        Model & \multicolumn{2}{c|}{Per-instance} & \multicolumn{2}{c}{Per-class} \\
        & Top-1 & Top-5 & Top-1 & Top-5 \\
        \bottomrule
        \rowcolor{aliceblue} Pose backbone & \textbf{62.6} & \textbf{81.2} & \textbf{41.9} & \textbf{63.2} \\ 
        - Contrastive loss & 61.7 & 80.5 & 41.0 & 62.1 \\ 
        \quad - Auxiliary loss & 60.8 & 80.0 & 39.8 & 60.9 \\ 
        \quad\quad - Feature masking & 59.5 & 78.8 & 37.8 & 58.8 \\ 
        \quad\quad\quad - Joint aggregation & 57.3 & 77.0 & 35.6 & 57.4 \\
        \midrule
        \rowcolor{aliceblue} Lip backbone & \textbf{34.8} & \textbf{58.8} & \textbf{19.2} & \textbf{39.4} \\ 
        - Contrastive loss & 33.9 & 57.9 & 18.6 & 38.8 \\ 
        \toprule
        \end{tabular}
    }
    \vspace{-0.4cm}
    \caption{\textbf{Visual backbone ablation.} We ablate key components for ISLR evaluation on the BOBSL-SIGN validation set.}
    \vspace{-0.3cm}
    \label{tab:islr_ablation}
\end{table}

\subsection{Ablation study}
\label{subsec:ablation}
\newpara{Visual backbone ablation.}
We conduct an ablation study on the visual backbones training and aggregation to evaluate the effectiveness of each component. 
For the pose backbone, we progressively remove (i) the contrastive loss, (ii) the auxiliary loss, (iii) the feature masking, and (iv) the joint aggregation. 
Here, the joint aggregation corresponds to the reshaping and projection operation described in \cref{subsec:architecture} (pose backbone), and its removal is implemented by replacing it with a simple mean pooling along the joint dimension. 
For the lip backbone, we ablate the contrastive loss. 
As shown in \cref{tab:islr_ablation}, removing any of these components consistently degrades performance on the BOBSL-SIGN (ISLR) validation set, confirming that all proposed components contribute positively to sign recognition.

\newpara{Mapping network.} 
To evaluate the effectiveness of the proposed Sliding Perceiver as a visual-to-text mapping network, we compare it with several alternatives: a standard MLP, a Conv1D + pooling-based network (both implemented following~\cite{jang2025lost}), and a Token Shuffle method~\cite{zeng2024timesuite}. 
In Token Shuffle, adjacent frames are stacked along the channel dimension and then projected back to the target embedding size in a manner similar to the Sliding Perceiver.
To isolate the effect of the mapping network, the visual backbone is frozen and all variants are trained for 10 epochs on the BOBSL-SENT training set with the SLT task, with evaluation on the BOBSL-SENT validation set. 
As shown in \cref{tab:abl_mapping}, the Sliding Perceiver outperforms all alternatives across all metrics, demonstrating its superior ability to aggregate local frame context while effectively bridging the visual and textual modalities.

\begin{table}
    \centering
    \setlength{\tabcolsep}{8pt}
    \resizebox{1\linewidth}{!}
    {
    \begin{tabular}{l|cccccc}
        \toprule
        Mapping  & B4 & B-RT & R-L & CIDEr & IoU & LLM \\
        \bottomrule
        MLP & 5.7 & 45.4 & 23.2 & 70.9 & 21.0 & 1.68 \\ 
        Conv-based & 5.9 & 46.7 & 24.0 & 74.5 & 21.8 & 1.70\\
        Token Shuffle~\cite{zeng2024timesuite} & 6.3 & 47.4 & 24.5 & 80.5 & 22.8 & 1.80 \\ 
        \rowcolor{aliceblue} Sliding Perceiver (Ours) & \textbf{6.6} & \textbf{47.6} & \textbf{25.6} & \textbf{82.5} & \textbf{23.2} & \textbf{1.85} \\ 
        \toprule
    \end{tabular}
    }
    \vspace{-0.4cm}
    \caption{\textbf{Mapping network.} We compare the proposed Sliding Perceiver with three alternatives -- a standard MLP, Conv-based mapping used in~\cite{jang2025lost}, and Token Shuffle~\cite{zeng2024timesuite} -- for SLT evaluation on the BOBSL-SENT validation set.}
    \vspace{-0.3cm}
    \label{tab:abl_mapping}
\end{table}

\begin{table}
    % \small
    \setlength{\tabcolsep}{9pt}
    \centering
    \resizebox{\linewidth}{!}
    {
        \begin{tabular}{l|cc|cc|c}
        \toprule
        Model & \multicolumn{2}{c|}{Per-instance} & \multicolumn{2}{c|}{Per-class} & Resource\\
        & Top-1 & Top-5 & Top-1 & Top-5 & GFLOPs\\
        \bottomrule
        Video-Swin-ISLR~\cite{raude2024} & 75.2 & 92.0 & 54.4 & 75.7 & 2.22\\ 
        \rowcolor{aliceblue} Pose backbone & 75.2 & 92.6 & 55.1 & 79.8 & 0.53\\ 
        \rowcolor{aliceblue} Lip backbone & 38.1 & 61.0 & 26.6 & 50.7 & 0.52\\ 
        \rowcolor{aliceblue} Ours (Pose+Lip) & \textbf{78.4} & \textbf{93.4} & \textbf{58.9} & \textbf{81.6} & 1.05\\ 
        \toprule
        \end{tabular}
    }
    \vspace{-0.4cm}
    \caption{\textbf{ISLR visual backbone performance (BOBSL).} The values are accuracies on the BOBSL-SIGN test set. GFLOPs indicate the computational cost to process a single frame.}
    % \az{maybe add number of params/memory here to explain lightweight} \jyj{training memory with 4 of batch size: } \jyj{GFLOPs per frame: Ours (pose): 0.53, Ours (lip): 0.52, Swin: 2.22 - lip reading model (0.495) conformer in lip backbone (0.25)}}
    \vspace{-0.3cm}
    \label{tab:islr_performance}
\end{table}
\subsection{Comparison to the state-of-the-art}
\label{subsec:sota}

\newpara{ISLR performance (BOBSL).}
We evaluate ISLR on the BOBSL-SIGN test set, comparing against the Video-Swin-ISLR~\cite{raude2024} baseline. 
As shown in \cref{tab:islr_performance}, the pose backbone alone achieves a Top-1 per-instance accuracy of 75.2\%, which matches the baseline, and exceeds the baseline on all other metrics.
The lip backbone performs worse -- as would be expected as signals such as mouthings are only occasional.
Late fusion of pose and lip logits (weight ratio 0.7:0.3, tuned on the validation set; see \cref{subsec:app:abl_fusion_ratio} of supp.\ mat.) yields the best model, demonstrating that the lip features are indeed complementary to the pose features. The fused model outperforms Video-Swin-ISLR across all metrics while requiring 1.05 GFLOPs, less than half of the 2.22 GFLOPs needed by the Swin model. 
We further provide a training memory analysis in \cref{subsec:app:efficiency} of the supp.\ mat.

\newpara{SLT performance (BOBSL).}
In \cref{tab:sota-bobsl}, we evaluate our model on SLT using the BOBSL-SENT test set. The \textit{Ours (Multitask)} model, which undergoes large-scale pretraining and is subsequently finetuned on BOBSL with a multitask objective (SLT + SSA), outperforms all prior approaches by a large margin. Notably, compared to LiTFiC (Full)~\cite{jang2025lost}, which relies on additional contextual inputs (e.g., previous subtitle sentence, pseudo-glosses, background captions) and a larger LLM (LLaMA3-8B), our model achieves more than double the BLEU-4 score using only video input.
The \textit{Ours (SLT)} model, starting from the same large-scale pretrained weights but finetuned on SLT only, achieves nearly identical performance, showing that multitask training with SSA does not compromise translation. For a fairer comparison, we also evaluate \textit{Ours (SLT, w/o PT)}, trained on SLT {\em without} large-scale pretraining. In this variant, the visual backbone (initialised with ISLR-pretrained weights), Sliding Perceiver, and DoRA~\cite{liu2024dora} adapters are trained for 10 epochs on BOBSL, following the same setup as~\cite{jang2025lost}. 
Even without large-scale pretraining, our model outperforms all prior works, showing that the visual backbone and Sliding Perceiver produce strong representations.

\begin{table}
    % \small
    \setlength{\tabcolsep}{6pt}
    \centering
    \resizebox{\linewidth}{!}
    {
    \begin{tabular}{l|cc|cccccc}
        \toprule
        Model & PT & hrs & B4 & B-RT & R-L & CIDEr & IoU & LLM \\
        \bottomrule
        Albanie~\cite{Albanie2021bobsl} & \xmark & - & 1.0 & - & 10.2 & - & - & - \\ 
        Sincan~\cite{sincan2023context} & \xmark & - & 1.3 & - & 8.9 & - & - & -\\
        GFSLT$^\dagger$~\cite{zhou2023gloss} & \xmark & - & 0.6 & 27.7 & 7.4 & 4.3 & 5.2 & 0.05 \\ 
        Sign2GPT$^\dagger$~\cite{wong2024sign2gpt} & \xmark & - & 0.9 & 35.2 & 11.4 & 16.1 & 8.7 & 0.49 \\ 
        LiTFiC (Vid)~\cite{jang2025lost} & \xmark & - & 2.6 & 37.8 & 15.6 & 37.5 & 13.6 & 0.95 \\ 
        LiTFiC (Full)~\cite{jang2025lost} & \xmark & - & 3.3 & 40.3 & 16.9 & 41.9 & 14.8 & 1.20 \\ 
        \rowcolor{aliceblue} Ours (SLT, w/o PT) & \xmark & - & 6.0 & 46.4 & 23.1 & 73.4 & 20.9 & 1.67 \\ 
        \rowcolor{aliceblue} Ours (SLT) & \cmark & 2.7K & 6.7 & \textbf{47.1} & \textbf{23.7} & 76.6 & \textbf{21.8} & 1.74 \\ 
        \rowcolor{aliceblue} Ours (Multitask) & \cmark & 2.7K & \textbf{6.8} & 47.0 & \textbf{23.7} & \textbf{77.0} & \textbf{21.8} & \textbf{1.77} \\ 
        \toprule         
    \end{tabular}
    }
    \vspace{-0.4cm}
    \caption{\textbf{SLT performance (BOBSL).} The results for GFSLT$^\dagger$ and Sign2GPT$^\dagger$ are sourced from \cite{jang2025lost}. Our model establishes a new state of the art on the BOBSL-SENT test set, significantly improving over prior results on several measures.
    % \zifan{In the figures comparing to previous, can we add their modalities used? I think it is clear that end2end pose+lip training is boosting our numbers greatly. While we avoided end2end training on full RGB pixels, which is too costly.}
    }
    \vspace{-0.3cm}
    \label{tab:sota-bobsl}
\end{table}

\begin{table}[t]
    \centering
    \setlength{\tabcolsep}{10pt}
    \resizebox{\linewidth}{!}{
    \begin{tabular}{l|ccccc}
        \toprule
        Model & Acc & CB-Acc & F1@.10 & F1@.25 & F1@.50 \\
        \bottomrule
        SAT~\cite{Bull21} & 71.0 & 57.3 & 71.0 & 74.2 & 53.4 \\
        SAT\textsuperscript{+}~\cite{jang2025deep} & \textbf{77.2} & 65.0 & 81.4 & 75.0 & 63.8 \\
        \rowcolor{aliceblue} Ours (SSA) & 75.7 & 66.9 & 83.7 & 79.7 & 67.0 \\
        \rowcolor{aliceblue} Ours (Multitask) & 77.1 & \textbf{67.5} & \textbf{84.4} & \textbf{80.3} & \textbf{68.9} \\
        \toprule
    \end{tabular}
    }
    \vspace{-0.4cm}
    \caption{\textbf{SSA performance (BOBSL).} 
    We evaluate alignment on the BOBSL-SENT test set. Our model not only achieves the best performance on the SSA task while performing SLT, but also shows a remarkable improvement in F1 scores. }
    \vspace{-0.3cm}
    \label{tab:ssa-sota-bobsl}
\end{table}

\newpara{SSA performance (BOBSL).}
In \cref{tab:ssa-sota-bobsl}, we compare our SSA performance to prior works, including SAT~\cite{Bull21} and SAT\textsuperscript{+}~\cite{jang2025deep}. All our models undergo large-scale pretraining and differ in finetuning: \textit{Ours (SSA)} is finetuned solely on the SSA task, while \textit{Ours (Multitask)} is finetuned jointly on SLT and SSA. Alignment supervision for both models uses pseudo-subtitles from SAT\textsuperscript{+}.
\textit{Ours (SSA)} achieves lower frame accuracy than SAT\textsuperscript{+} (-1.5\%) but improves F1@0.5 by 3.2. \textit{Ours (Multitask)} attains comparable frame accuracy to SAT\textsuperscript{+}, while CB-Acc and F1@0.5 increase by 2.5\% and 5.1, respectively. These results show that multitask training improves alignment, yielding better signing-aligned subtitles than the pseudo-labels used for supervision.

\newpara{Zero-shot SLT performance (How2Sign, FLEURS-ASL).}
In \cref{tab:zero-shot}, we  evaluate the large-scale pretrained model in a zero-shot setting on two ASL datasets, How2Sign and FLEURS-ASL. Here, \emph{zero-shot} refers to neither of the datasets having been seen at any point during training -- note however that the ASL language is present in YouTube-SL-25. 
On the How2Sign, our model achieves a B-RT score of 51.3, surpassing all previously reported zero-shot performances. 
Remarkably, this result is comparable to Uni-Sign~\cite{li2025uni} and Geo-Sign~\cite{fish2025geo}, which are pretrained on 1,985 hours of Chinese Sign Language data and finetuned on the How2Sign (see \cref{tab:sota-how2sign}).
Similarly, on the FLEURS-ASL, our model also surpasses the previous state-of-the-art FLEURS-SLT~\cite{tanzer2024fleurs} model. These findings further highlight the general representations learned by our model, which captures visual–textual representations transferable across sign languages, linguistic domains, and data distributions. 

\newpara{SLT performance (How2Sign).}
Here we finetune our large-scale pretrained model on How2Sign. For this dataset, we train the model solely on the SLT task, as we observe that the temporal offset between audio-aligned and signing-aligned subtitles frequently exceeds our 20-second temporal window (at 25 fps), making SSA supervision unreliable.
Evaluation results on the How2Sign test set are given in \cref{tab:sota-how2sign}.
Our model (\textit{Ours}) surpasses all previous methods by a substantial margin, including SSLT~\cite{zhang2024scaling}, which is pretrained on a much larger corpus of 6,600 hours of sign videos compared to ours (2,700 hours).
For reference, a {\em human} baseline~\cite{tanzer2024fleurs} on a small subset of the test set (so, not fully representative)  reports B4: 19.8 and B-RT: 56.6, a similar performance to our result.
Even without the pretraining, this model outperforms VAP~\cite{jiao2024visual} and LiTFiC~\cite{jang2025lost}, which do not leverage extra pretraining data,
and achieves results comparable to models pretrained on over 1K hours of sign videos, including SSVP-SLT~\cite{rust2024towards}, Uni-Sign~\cite{li2025uni}, and Geo-Sign~\cite{fish2025geo}.
These results highlight the strong generalisation capability of our model.

\begin{table}
    % \small
    \setlength{\tabcolsep}{9pt}
    \centering
    \resizebox{1\linewidth}{!}
    {
        \begin{tabular}{l|cccccc}
        \toprule
        Model  & B4 & B-RT & R-L & CIDEr & IoU & LLM \\
        \bottomrule
        \rowcolor{gray!10} \multicolumn{7}{c}{\textsc{How2Sign}} \\
        FLEURS-SLT~\cite{tanzer2024fleurs} & 5.1 & 36.9 & - & - & - & -\\ 
        SSLT~\cite{zhang2024scaling} & 5.8 & 37.1 & - & - & - & -\\ 
        SSVP-SLT~\cite{rust2024towards} & 7.1 & 41.8 & 28.3 & - & - & - \\ 
        \rowcolor{aliceblue} Ours & \textbf{13.6} & \textbf{51.3} & \textbf{35.7} & \textbf{123.2}  & \textbf{30.8} & \textbf{2.10} \\
        \bottomrule
        \rowcolor{gray!10} \multicolumn{7}{c}{\textsc{FLEURS-ASL}} \\
        FLEURS-SLT~\cite{tanzer2024fleurs} & 3.9 & 38.3 & - & - & - & -\\ 
        % SSLT~\cite{zhang2024scaling} & 4.4 & 40.1 & - & - & - & -\\ 
        \rowcolor{aliceblue} Ours & \textbf{13.0} & \textbf{59.5} & \textbf{37.7} & \textbf{131.9}  & \textbf{34.5} & \textbf{2.98} \\
        \toprule
        \end{tabular}
    }
    \vspace{-0.4cm}
    \caption{
    \textbf{Zero-shot SLT performance.} Results on the How2Sign test and FLEURS-ASL devtest sets. Our model shows significant gains over prior works, demonstrating strong generalisation.
    }
    \vspace{-0.3cm}
    \label{tab:zero-shot}
\end{table}
\begin{table}
    \centering
    \resizebox{1\linewidth}{!}{
        \begin{tabular}{l|cc|cccccc}
        \toprule
        Model  & PT & hrs & B4 & B-RT & R-L & CIDEr & IoU & LLM \\
        \bottomrule
        VAP~\cite{jiao2024visual} & \xmark & - & 12.9 & - & 27.8 & - & - & - \\ 
        LiTFiC~\cite{jang2025lost} & \xmark & - & 12.7 & 45.3 & 32.5 & 100.8  & 27.9 & 1.59 \\
        SSLT~\cite{zhang2024scaling} & \cmark & 6.6k & 21.1 & 55.7 & - & - & - & -\\ 
        SSVP-SLT~\cite{rust2024towards} & \cmark & 1.0k & 15.5 & 49.6 & 38.4 & - & - & - \\ 
        Uni-Sign~\cite{li2025uni} & \cmark & 1.9k & 14.9 & 49.4 & 36.0 & - & - & - \\ 
        Geo-Sign~\cite{fish2025geo} & \cmark & 1.9k & 15.1 & - & 35.4 & - & - & - \\ 
        \rowcolor{aliceblue} Ours (w/o PT) & \xmark & - & 17.5 & 49.5 & 38.4 & 147.5 & 32.2 & 1.97 \\
        \rowcolor{aliceblue} Ours & \cmark & 2.7k & \textbf{24.6} & \textbf{58.7} & \textbf{48.1} & \textbf{219.0} & \textbf{43.7} & \textbf{2.92} \\
        % \rowcolor{aliceblue} Ours & \cmark & 2.7k & \textbf{24.4} & \textbf{58.2} & \textbf{47.6} & \textbf{218.3} & \textbf{43.6} & \textbf{2.92} \\
        \toprule
        \end{tabular}
    }
    \vspace{-0.4cm}
    \caption{
    \textbf{SLT performance (How2Sign).} We report results on the test set with/without large-scale pretraining (PT), highlighting its importance. Our final model achieves the best performance.
    }
    \vspace{-0.3cm}
    \label{tab:sota-how2sign}
\end{table}

\section{Conclusion}
\label{sec:conclusion}
In this work, we leverage both pose and lip features to build generalisable embeddings for sign language understanding, achieving state-of-the-art performance on ISLR, SLT, and SSA tasks. Lip features complement pose by capturing fine-grained mouthing details, while using both modalities enables the design of a lightweight visual backbone that can be trained end-to-end with large-scale datasets and LLMs. Through extensive experiments, we demonstrate that these representations are highly transferable across languages, datasets, and signers.
Despite our improvements, using 20-second signing windows poses limitations, especially when subtitles or the lag between audio and signing aligned subtitles exceeds this duration. Tackling these challenges could enhance both the reliability and practical deployment of sign language accessibility technologies.

{
\paragraph{Acknowledgments.}
The images in this paper are used with kind permission of the BBC.
This work was supported by the ANR project
CorVis ANR-21-CE23-0003-01, the UKRI EPSRC Programme Grant SignGPT EP/Z535370/1, and a Royal Society Research Professorship RSRP$\backslash$R$\backslash$241003,
the Institute of Information \& communications Technology Planning \& Evaluation (IITP) grant funded by the Korean government (MSIT, RS-2025-02263977, Development of Communication Platform supporting User Anonymization and Finger Spelling-Based Input Interface for Protecting the Privacy of Deaf Individuals). ZJ is funded by the Swiss Innovation Agency (Innosuisse) flagship IICT (PFFS-21-47).
}

{
    \small
    \bibliographystyle{ieeenat_fullname}
    \bibliography{shorstrings, main, vgg_local}
}
\clearpage
\maketitlesupplementary

\appendix

\renewcommand{\thefigure}{A.\arabic{figure}} 
\setcounter{figure}{0} 
\renewcommand{\thetable}{A.\arabic{table}}
\setcounter{table}{0} 

\startcontents[sections]
{
	\hypersetup{linkcolor=black}
	\printcontents[sections]{l}{1}{}
}

\vspace{10mm}

\section{Implementation details}

\subsection{Pose backbone encoder}
\label{subsec:app:pose_backbone}
\begin{figure*}[t]
  \centering
  \includegraphics[width=\linewidth]{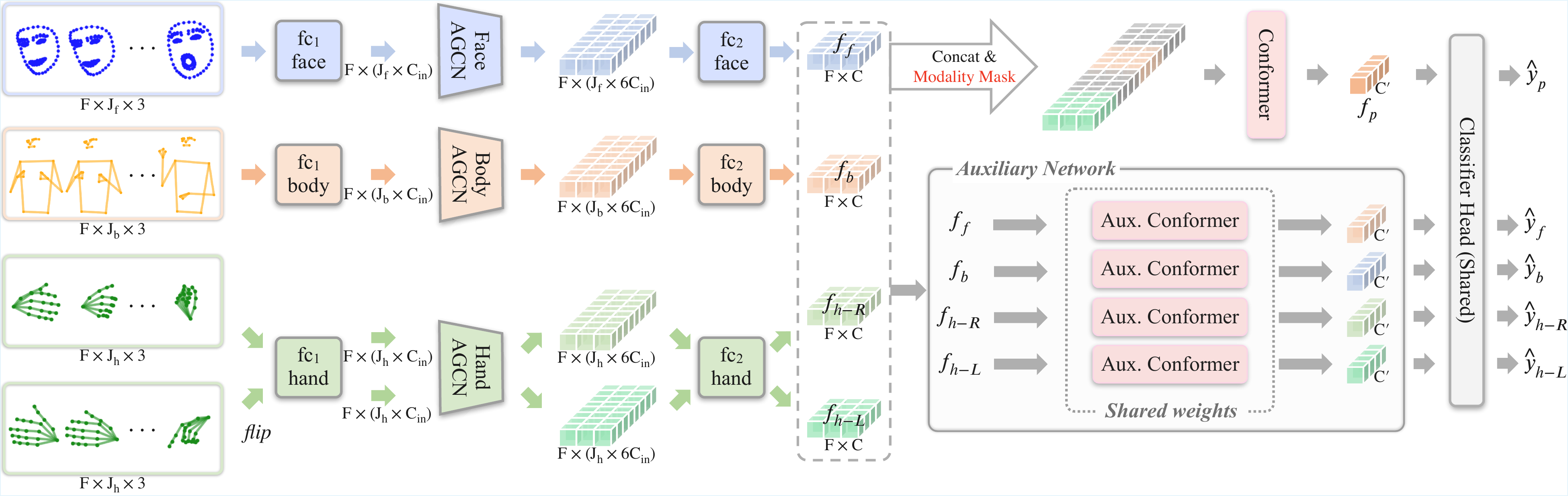}
  \caption{\textbf{A detailed pose backbone illustration.} \textbf{\textit{(i)~Main path}:} Starting from the 3D keypoints, each articulator stream processes its input through an initial fully connected layer ($\mathrm{fc}_{1}$) followed by a 4-layer AGCN. The resulting jointwise features are reshaped and aggregated using $\mathrm{fc}_{2}$ to obtain the articulator-specific representations ($f_f$, $f_b$, $f_{h\text{-}R}$, $f_{h\text{-}L}$). During training, we randomly zero-fill up to two of these four features to mask articulators, after which the features pass through the backbone Conformer and the classifier head to produce the pose logits $\hat{y}_{p}$.
  \textbf{\textit{(ii)~Auxiliary network}:}
  The articulator-specific features ($f_f$, $f_b$, $f_{h\text{-}R}$, $f_{h\text{-}L}$) are each processed by a shared-weight auxiliary Conformer, and the resulting features are subsequently passed to the same classifier heads employed in the pose backbone. The classifier produces articulator-specific logits ($\hat{y}_f$, $\hat{y}_b$, $\hat{y}_{h\text{-}R}$, $\hat{y}_{h\text{-}L}$), which are compared against the pseudo-gloss label $y$ using a cross-entropy objective, providing additional supervision during training.
  }
  \label{fig:app:pose_backbone_full}
\end{figure*}

\cref{tab:app:vis_backbone} reports the architecture details and input--output dimensions of the articulator-specific encoder in our pose backbone.
Here, $F$ denotes the number of input frames, $J_{m}$ the number of joints for each articulator, $C_{\text{in}}$ the input channels to the 4-layer AGCN, and $C_{out}$ the output channels of the articulator-specific encoders.
Instead of applying mean pooling across the joint dimension, we aggregate joints by reshaping followed by a fully connected layer $\mathrm{fc}_{2}$, reducing information loss during joint fusion. A detailed illustration of the full forward pass of the pose backbone is provided in \cref{fig:app:pose_backbone_full}.

\subsection{Conformer}
\label{subsec:app:conformer}
\cref{tab:app:conformer} presents the architectural details of the Conformer modules used in the pose backbone, the lip backbone, and the auxiliary network jointly trained with the pose backbone. Each Conformer maps input features of dimension $C_{\text{in}}$ to a hidden size $C'$, processes them through four Conformer blocks, and applies an attentive pooling layer for temporal aggregation. 

\begin{table}
    \setlength{\tabcolsep}{12pt}
	\centering
	\resizebox{\columnwidth}{!}{
		\begin{tabular}{c|c|c}
			Layer & Input sizes & Output sizes \\
			\hline
			{fc$_1$} & $F \times J_{m} \times 3$ & $F \times J_{m} \times C_{in}$ \\
			\hline
			{AGCN$_1$} & $F \times J_{m} \times C_{in}$ & $F \times J_{m} \times 2C_{in}$ \\
			\hline
			{AGCN$_2$} & $F \times J_{m} \times 2C_{in}$ & $F \times J_{m} \times 2C_{in}$ \\
			\hline
			{AGCN$_3$} & $F \times J_{m} \times 2C_{in}$ & $F \times J_{m} \times 3C_{in}$ \\
			\hline
			{AGCN$_4$} & $F \times J_{m} \times 3C_{in}$ & $F \times J_{m} \times 6C_{in}$ \\
			\hline
			{reshaping} & $F \times J_{m} \times 6C_{in}$ & $F \times (J_{m} \times 6C_{in})$ \\
			\hline
			{fc$_2$} & $F \times (J_{m} \times 6C_{in}$) & $F \times C_{out}$ \\
	\end{tabular}
        }
	\caption{\textbf{Articulator-specific encoder in our pose backbone.}
    We set $F=24$. The joint counts $J_{m}$ vary by articulator: $128$ for the face, $33$ for the body, and $21$ each for the left and right hands. $C_{\text{in}}=64$, and the encoder outputs $C_{out}=384$ channels.
	} 
	\label{tab:app:vis_backbone}
\end{table}
\begin{table}
    \setlength{\tabcolsep}{14pt}
	\centering
	\resizebox{\columnwidth}{!}{
		\begin{tabular}{c|c|c}
			Layer & Input sizes & Output sizes \\
			\hline
			{fc$_1$} & $F \times C_{in}$ & $F \times C'$ \\
			\hline
			{Conformer block$_1$} & $F \times C'$ & $F \times C'$ \\
			\hline
			{Conformer block$_2$} & $F \times C'$ & $F \times C'$ \\
			\hline
			{Conformer block$_3$} & $F \times C'$ & $F \times C'$ \\
			\hline
			{Conformer block$_4$} & $F \times C'$ & $F \times C'$ \\
			\hline
			{Attentive pool} & $F \times C'$ & $C'$ \\

	\end{tabular}
        }
	\caption{\textbf{Conformer.} The input dimensions are $C_{\text{in}} = 1{,}536$ for the pose backbone, $768$ for the lip backbone, and $384$ for the auxiliary Conformer, while all three use a shared hidden size of $C' = 512$.
	}
	\label{tab:app:conformer}
\end{table}
\begin{table}
    \setlength{\tabcolsep}{24pt}
	\centering
	\resizebox{\columnwidth}{!}{
		\begin{tabular}{c|c|c}
			Layer & Input sizes & Output sizes \\
			\hline
			{fc$_1$} & $C_{in}$ & $C'$ \\
			\hline
			{ReLU} & $C'$ & $C'$ \\
			\hline
			{fc$_2$} & $C'$ & $C_{out}$ \\
	\end{tabular}
        }
	\caption{\textbf{MLP.} The visual features $f_p$ and $f_l$ both have $512$ channels; hence the visual projector uses $C_{\text{in}} = 512$, with a hidden size of $C' = 256$ and output dimension of $C_{\text{out}} = 128$.  For the text stream, the projector takes the $2{,}048$-dimensional LLM embedding as input ($C_{\text{in}} = 2{,}048$), maps it through a hidden size of $C' = 512$, and outputs a $128$-dimensional representation ($C_{\text{out}} = 128$). 
	}
	\label{tab:app:mlp}
\end{table}

\begin{figure}[t]
  \centering
  \includegraphics[width=\linewidth]{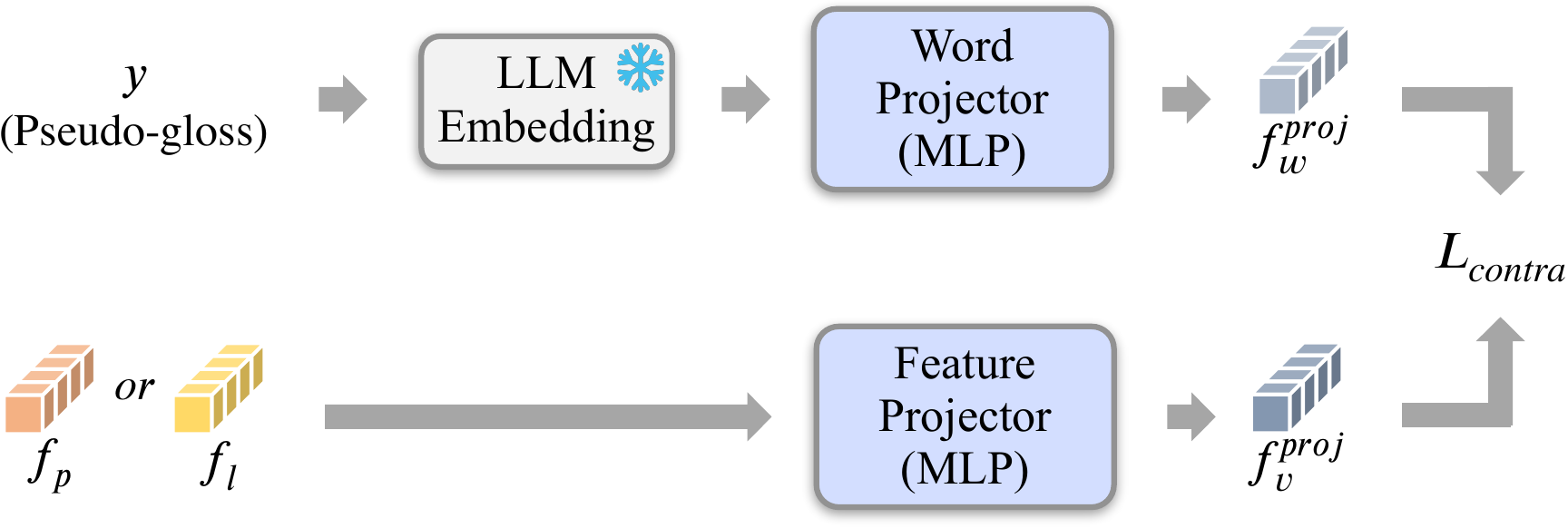}
  \caption{\textbf{ISLR-pretraining with contrastive loss.} The pseudo-gloss $y$ is passed sequentially through a frozen LLM embedding and an MLP projector to obtain $f_w^{proj}$. The visual features ($f_p$ or $f_l$) are passed through the MLP projector to obtain $f_v^{proj}$. Finally, $f_w^{proj}$ and $f_v^{proj}$ are used to compute the contrastive loss.}
  \label{fig:app:contrastive_loss}
\end{figure}

\subsection{MLP projector for constrastive loss}
\label{subsec:app:mlp}
As explained in \appendixref{Sec.~3.3}{\cref{subsec:backbone}} of the main paper, we employ an MLP-based projector to align the visual and textual embeddings in a shared space with a contrastive loss during ISLR pretraining (see \cref{fig:app:contrastive_loss} for an illustration). The projector consists of two fully connected layers with a \texttt{ReLU} activation, and its architectural details are provided in \cref{tab:app:mlp}.

\subsection{Auxiliary network overview}
As discussed in \appendixref{Sec.~3.3}{\cref{subsec:backbone}} of the main paper, we reinforce the pose backbone representation using an auxiliary articulator loss. The architecture is illustrated in the bottom-right of \cref{fig:app:pose_backbone_full}.

\subsection{Sliding Perceiver}
\begin{figure}[t]
  \centering
  \includegraphics[width=\linewidth]{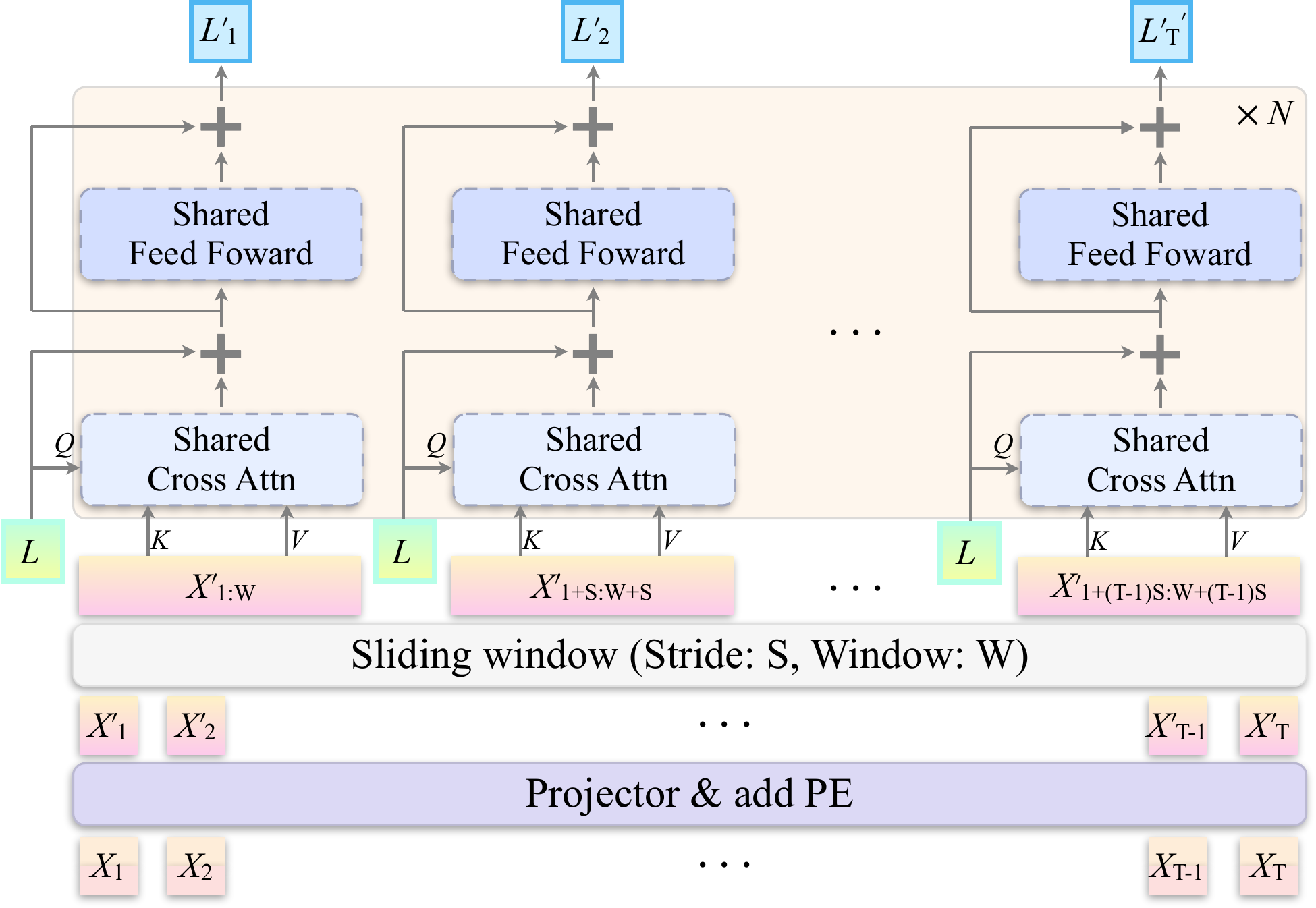}
  \caption{\textbf{Sliding Perceiver.} The input \(X\) is projected to the LLM dimension \(D = 2{,}048\) via a projector, after which global positional encoding is added to obtain \(X'\). A sliding window (stride 2, window size 8) is then applied to \(X'\) to form local features, which interact with the latent \(L\) through \(N = 2\) stacked cross-attention and feed-forward layers to produce \(L'\). In the cross-attention blocks, \(L\) serves as the queries, while the locally extracted features provide the keys and values.
  }
  \label{fig:app:sliding_perceiver}
\end{figure}

Supplementing the description of our Sliding Perceiver in \appendixref{Sec.~3.1}{\cref{subsec:architecture}}, \cref{fig:app:sliding_perceiver} illustrates, at the token-level, how the Sliding Perceiver produces $L'$.

\begin{figure*}[t]
  \centering
  \includegraphics[width=\linewidth]{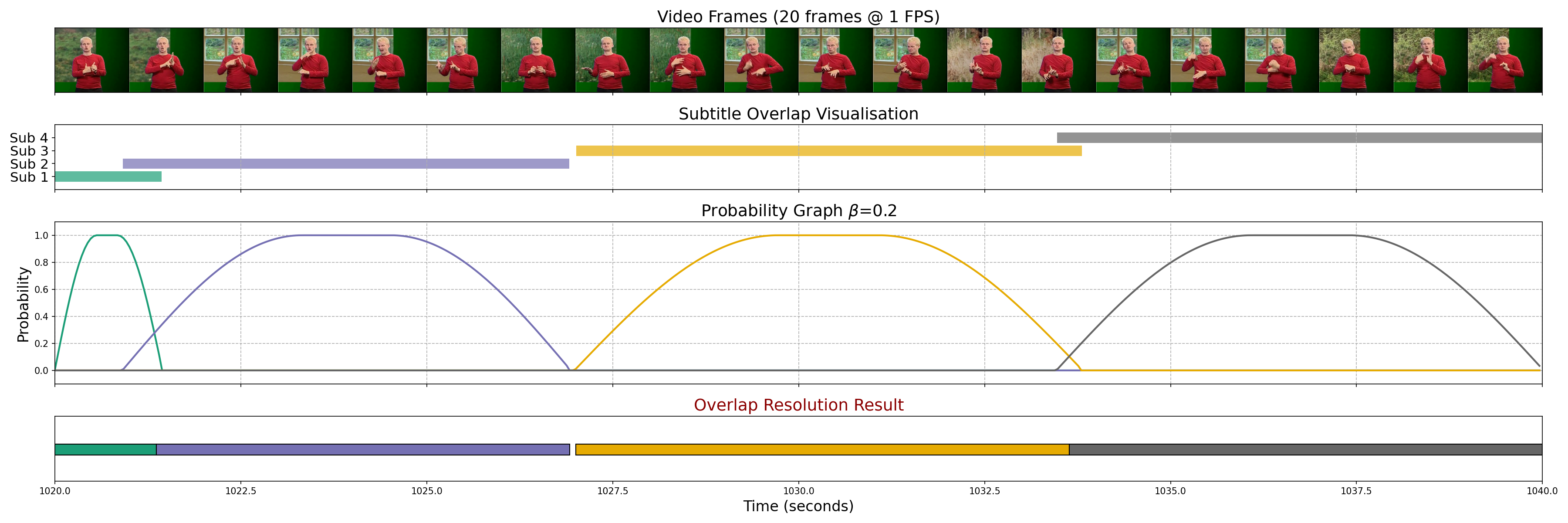}
  \caption{\textbf{Subtitle postprocessing.} From top to bottom, the panels display: sampled video frames for context; the initial predicted timestamps containing temporal overlaps; the generated probability profiles used as soft constraints; and the final resolved intervals. The probability maps guide the alignment algorithm to determine optimal boundaries, preventing the suppression of shorter sequences.}
  \label{fig:app:remove_overlap} 
\end{figure*}

\begin{table}[t]
    \centering
    \setlength{\tabcolsep}{12pt}
    \resizebox{\linewidth}{!}{
    \begin{tabular}{l|ccccc}
        \toprule
        $\beta$ & Acc & CB-Acc & F1@.10 & F1@.25 & F1@.50 \\
        \bottomrule
        0.0 & 85.4 & 78.9 & 91.3 & 88.7 & 80.5 \\
        0.1 & 85.4 & \textbf{79.0} & 91.4 & \textbf{88.8} & 80.5 \\
        \rowcolor{aliceblue} 0.2 & \textbf{85.5} & \textbf{79.0} & \textbf{91.5} & \textbf{88.8} & \textbf{80.6} \\
        0.3 & 85.4 & \textbf{79.0} & 91.3 & 88.6 & 80.5 \\
        0.4 & 85.4 & 78.7 & 91.1 & 88.3 & 80.4 \\
        0.5 & 85.3 & 78.5 & 90.9 & 88.0 & 80.0 \\
        \toprule
    \end{tabular}
    }
    \caption{\textbf{Ablation on DTW hyperparmeter $\beta$ on the \textsc{BOBSL-Sent} validation set.} The model achieves the best alignment performance when $\beta = 0.2$.
    }
    \label{tab:app:dtw}
\end{table}

\subsection{DTW subtitle postprocessing for SSA}
\label{subsec:app:dtw}
As described in \appendixref{Sec.~4.2}{\cref{subsec:implementation}}, unlike previous works~\cite{Bull21,jang2025deep} relying on frame-level binary classification, our model utilises an LLM to predict start and end timestamps. To resolve potential overlaps between adjacent subtitles—where a simple equal split would disproportionately truncate shorter sequences—we construct a temporal probability map to serve as input for DTW, as illustrated in \cref{fig:app:remove_overlap}.
Specifically, we model the probability profile of each subtitle using a plateau-decay function. The central region of the subtitle retains a probability of 1.0 to preserve the core content. The duration of this plateau is controlled by a hyperparameter $\beta$, representing the fraction of the total subtitle duration (fixed at $\beta=0.2$ in our experiments, meaning the central 20\% is preserved).
For the remaining duration (the fade-in and fade-out regions), the probability follows a cosine curve to ensure a smooth transition. The rising edge is modelled by $\cos(\theta)$ where $\theta$ maps from $[-\frac{\pi}{2}, 0]$, and the falling edge maps from $[0, \frac{\pi}{2}]$. This soft probability map is then used to determine the optimal boundary between overlapping subtitles, prioritising the retention of shorter sequences while resolving conflicts. The alignment performance across different values of $\beta$ is reported in \cref{tab:app:dtw}

\subsection{Text cleaning procedures for YouTube-SL-25}
\label{subsec:app:text_cleaning}
As mentioned in \appendixref{Sec.~4.1}{\cref{subsec:data}}, to remove noise from the raw subtitle texts of~\cite{tanzer2024youtube}, we apply a multi-stage cleaning pipeline. 
We first decode HTML entities (e.g., \texttt{\&lt;} $\rightarrow$ \textless) and remove any remaining HTML tags such as \texttt{<font>} or \texttt{<c>}. 
Next, we eliminate non-verbal annotations or stylistic markers enclosed in parentheses, brackets, or asterisks (e.g., \texttt{(laughter)}, \texttt{[music]}, \texttt{*flush*}). 
We then remove leading special symbols (e.g., `♪', `•', `-') and discard speaker identifiers when a single word followed by a colon appears at the beginning of a line (e.g., \texttt{John:}). 
To ensure textual consistency, we also delete unnecessary spaces before punctuation marks, collapse consecutive spaces into one, and trim leading or trailing quotation marks. 

\subsection{Class-balanced accuracy (CB-Acc)}
\label{subsec:app:cb_acc}
As described in \appendixref{Sec.~4.1}{\cref{subsec:data}} of the main paper, we introduce class-balanced accuracy (CB-Acc) to evaluate temporal alignment performance, addressing imbalances due to varying subtitle lengths and non-signing periods.

For a video clip with $N_\text{sub}$ subtitles, we define the classes such that background frames (no signing) are assigned to class 0, and subtitle frames are assigned class indices 1 through $N_\text{sub}$. CB-Acc is computed using the standard balanced accuracy formulation. Specifically, we first construct a confusion matrix $C$ between the ground-truth labels $y_\text{true}$ and predicted labels $y_\text{pred}$. For each class $i \in \{0, \dots, N_\text{sub}\}$, the recall is computed as
\[
\text{recall}_i = \frac{C_{i,i}}{\sum_j C_{i,j}},
\]
where $C_{i,i}$ is the number of correctly predicted frames for class $i$, and $\sum_j C_{i,j}$ is the total number of ground-truth frames of class $i$. The CB-Acc is then obtained by averaging the per-class recalls:
\[
\text{CB-Acc} = \frac{1}{N_\text{sub}+1} \sum_{i=0}^{N_\text{sub}} \text{recall}_i.
\]

This formulation ensures that each subtitle and the non-signing frames contribute equally to the overall accuracy, regardless of their duration in the video.

\begin{table}
    % \small
    \setlength{\tabcolsep}{12pt}
    \centering
    \resizebox{\linewidth}{!}
    {
        \begin{tabular}{cc|cc|cc}
        \toprule
        \multicolumn{2}{c|}{Fusion ratio} & \multicolumn{2}{c|}{Per-instance} & \multicolumn{2}{c}{Per-class}\\
        pose & lip & Top-1 & Top-5 & Top-1 & Top-5\\
        \midrule
        0.5 & 0.5 & 64.9 & 83.1 & 43.5 & 63.9 \\ 
        0.6 & 0.4 & 66.1 & \textbf{83.8} & \textbf{45.3} & 63.9 \\ 
        \rowcolor{aliceblue} 0.7 & 0.3 & \textbf{66.2} & 83.7 & \textbf{45.3} & \textbf{65.9} \\ 
        0.8 & 0.2 & 65.5 & 83.2 & 44.6 & 65.8\\ 
        0.9 & 0.1 & 64.2 & 82.3 & 43.3 & 64.5\\ 
        \bottomrule
        \end{tabular}
    }
    \caption{\textbf{Ablation on fusion ratio for ISLR on \textsc{BOBSL-Sign} validation set.} The best performance is achieved when the logits from the pose backbone and lip backbone are fused with a $0.7:0.3$ ratio.}
    \label{tab:app:fusion}
\end{table}

\section{Additional experiments and analyses}

\subsection{Ablation on late fusion ratio of visual backbone}
\label{subsec:app:abl_fusion_ratio}
As mentioned in \appendixref{Sec.~4.4}{\cref{subsec:ablation}}, we perform late fusion by combining the logits from the pose backbone and the lip backbone with weights of $0.7$ and $0.3$, respectively. 
The ablation study for this fusion ratio is provided in \cref{tab:app:fusion}, where the $0.7:0.3$ configuration yields the best performance overall.

\subsection{Effectiveness of soft-decoding}
As introduced in \appendixref{Sec.~3.4}{\cref{subsec:downstream}}, we employ a soft-decoding approach to encourage predicted timestamps to be closer to the target timestamps. 
\cref{tab:app:soft-decoding} presents the effect of this technique on the \textsc{BOBSL-Sent} validation set. 
The results demonstrate that soft-decoding consistently improves performance across all evaluation metrics.

\subsection{Analysis on efficiency of visual backbone}
\label{subsec:app:efficiency}
To complement the comparison of Ours with Video-Swin-ISLR~\cite{raude2024} in \appendixref{Sec.~4.4}{\cref{subsec:sota}} of the main paper, we compare the training memory requirements of our visual backbone with those of the Video-Swin-ISLR model. When processing 500-frame inputs with stride 2 for dense feature extraction and batch size 1, Video-Swin-ISLR (window size 16) exceeds 96 GB of GPU memory, whereas our model (window size 24), using pre-extracted lip-reading features as input, requires only 29.6 GB. Notably, despite using a larger window size, our model consumes substantially less memory, highlighting its efficiency. This indicates that our model can be trained jointly with the LLM on a 48 GB GPU with batch size 1, while Video-Swin-ISLR cannot.

Despite being feasible, training with our model in this setting would remain extremely time-consuming, and the maximum batch size is constrained to 1. On an A6000 GPU, 400K iterations would take approximately 280 hours.

To further reduce memory usage, we modified the pipeline so that the AGCN-based encoder in the backbone processes all 500 frames at once (without sliding), while only the Conformer applies sliding-window feature extraction, leveraging its prior knowledge learned from 24-frame windows. With this approach, GPU memory usage drops to 5.6 GB (batch size 1), allowing an increase of batch size to 4 and reducing training time to roughly 68 hours per 100 K iterations.

\subsection{Sentence-level BLEURT comparison with LiTFiC}
To analyse the performance difference between our model and the previous state-of-the-art LiTFiC~\cite{jang2025lost} on the BOBSL dataset, we compare BLEURT scores across all 20,338 sentences in the \textsc{BOBSL-Sent} test set. \cref{tab:app:bobsl_win} shows the number of samples where each model outperforms the other. Our model achieves higher BLEURT scores on 13,647 samples (67.1\% of the total), demonstrating its overall superior performance in sign language translation.
\begin{table}[t]
    \centering
    \setlength{\tabcolsep}{5pt}
    \resizebox{\linewidth}{!}{
    \begin{tabular}{l|ccccc}
        \toprule
        Model & Acc & CB-Acc & F1@.10 & F1@.25 & F1@.50 \\
        \bottomrule
        Ours w/o soft-decoding & 84.6 & 78.1 & 90.8 & 88.0 & 79.7 \\
        \rowcolor{aliceblue} Ours & \textbf{85.5} & \textbf{79.0} & \textbf{91.5} & \textbf{88.8} & \textbf{80.6}\\ \toprule
    \end{tabular}
    }
    \caption{\textbf{Effectiveness of soft-decoding.} When SSA training uses soft-decoding, combining cross-entropy loss with an L1 loss on the soft-decoded and ground-truth timestamps leads to improvements on all metrics.}
    \label{tab:app:soft-decoding}
\end{table}

\begin{table}[t]
    \centering
    \setlength{\tabcolsep}{14pt}
    \resizebox{\linewidth}{!}{
    \begin{tabular}{c|c|c}
        \toprule
        LiTFiC~\cite{jang2025lost} \#win & Ours \#win & \#Total\\
        \bottomrule
        $6{,}691$ ($32.9\%$) & $13{,}647$ ($67.1\%$) & $20{,}338$ ($100\%$)\\\toprule
    \end{tabular}
    }
    \caption{\textbf{Comparison of SLT performance with LiTFiC on the \textsc{BOBSL-Sent} test set.} We report the number of samples where each model achieves a higher BLEURT score. Our model wins on $67.1\%$ of the $20{,}338$ test samples.}
    \label{tab:app:bobsl_win}
\end{table}

\begin{table*}[t]
    \centering
    \setlength{\tabcolsep}{8pt}
    \renewcommand{\arraystretch}{1.2}
    \resizebox{\linewidth}{!}{
    \begin{tabular}{c|c|c|c|c|c|c|c|c|c|c|c|c|c|c|c} % 13 columns: 4 + 9
        \toprule
        \multirow{2}{*}{\textbf{No.}} & \multicolumn{4}{c|}{\textbf{LLM Component}} &
        \multirow{2}{*}{\textbf{Decoding}} &
        \multirow{2}{*}{\textbf{Alignment}} &   \multicolumn{2}{c|}{\textbf{Vis. backbone}} &
        \multirow{2}{*}{\textbf{Mapping}} &
        \multicolumn{6}{c}{\textbf{Metrics}} \\
        \cmidrule(lr){2-5}\cmidrule(lr){8-9}\cmidrule(lr){11-16}
        & \textbf{LLM} & \textbf{FT} & \textbf{r} & \textbf{a} 
        & & & \textbf{Type} & \textbf{Update} & & \textbf{B4} & \textbf{B-RT} & \textbf{R-L} & \textbf{CIDEr} & \textbf{IoU} & \textbf{LLM} \\
        \midrule
        \rowcolor{gray!10} \multicolumn{16}{c}{\textsc{Baseline}~\cite{jang2025lost}} \\
        (1) & LlaMA-8B~\cite{touvron2023llama} & LoRA~\cite{hu2022lora} & 4 & 16 & greedy & SAT~\cite{Bull21} & Swin~\cite{raude2024} & \xmark & MLP & 3.4 & 41.0 & 18.6 & 48.9 & 16.6 & 1.29 \\
        \rowcolor{gray!10} \multicolumn{16}{c}{\textsc{LLM Component Ablation}} \\
        (2) & FLAN-T5-XL~\cite{chung2024scaling} & LoRA~\cite{hu2022lora} & 4 & 16 & greedy & SAT~\cite{Bull21} & Swin~\cite{raude2024} & \xmark & MLP & 3.7 & 42.0 & 19.2 & 51.0 & 17.2 & 1.37 \\
        (3) & FLAN-T5-XL~\cite{chung2024scaling} & DoRA~\cite{liu2024dora} & 4 & 16 & greedy & SAT~\cite{Bull21} & Swin~\cite{raude2024} & \xmark & MLP & 3.9 & 42.4 & 19.5 & 51.7 & 17.5 & 1.41 \\
        (4) & FLAN-T5-XL~\cite{chung2024scaling} & DoRA~\cite{liu2024dora} & 32 & 64 & greedy & SAT~\cite{Bull21} & Swin~\cite{raude2024} & \xmark & MLP & 4.6 & 43.0 & 20.8 & 56.8 & 18.1 & 1.50 \\
        \rowcolor{gray!10} \multicolumn{16}{c}{\textsc{Alignment Ablation}} \\
        (5) & FLAN-T5-XL~\cite{chung2024scaling} & DoRA~\cite{liu2024dora} & 32 & 64 & greedy & SAT\textsuperscript{+}\cite{jang2025deep} & Swin~\cite{raude2024} & \xmark & MLP & 4.8 & 43.3 & 21.0 & 58.0 & 18.4 & 1.52 \\
        \rowcolor{gray!10} \multicolumn{16}{c}{\textsc{Feature Ablation}} \\
        (6) & FLAN-T5-XL~\cite{chung2024scaling} &         DoRA~\cite{liu2024dora} & 32 & 64 & greedy & SAT\textsuperscript{+}\cite{jang2025deep} & Our vis. backbone & \xmark & MLP & 5.3 & 44.6 & 22.1 & 65.6 & 19.9 & 1.61 \\
        (7) & FLAN-T5-XL~\cite{chung2024scaling} &         DoRA~\cite{liu2024dora} & 32 & 64 & greedy & SAT\textsuperscript{+}\cite{jang2025deep} & Our vis. backbone (20s) & \xmark & MLP & 5.7 & 45.4 & 23.2 & 70.9 & 21.0 & 1.68 \\
        \rowcolor{gray!10} \multicolumn{16}{c}{\textsc{Mapping Network Ablation}} \\
        (8) & FLAN-T5-XL~\cite{chung2024scaling} &         DoRA~\cite{liu2024dora} & 32 & 64 & greedy & SAT\textsuperscript{+}\cite{jang2025deep} & Our vis. backbone (20s) & \xmark & Sliding Perceiver & 6.6 & 47.6 & 25.6 & 82.5 & 23.2 & 1.85 \\
        \rowcolor{gray!10} \multicolumn{16}{c}{\textsc{End-to-End Training}} \\
        (9) & FLAN-T5-XL~\cite{chung2024scaling} &         DoRA~\cite{liu2024dora} & 32 & 64 & greedy & SAT\textsuperscript{+}\cite{jang2025deep} & Our vis. backbone (20s) & \cmark & Sliding Perceiver & 7.4 & 48.6 & 26.6 & 90.2 & 23.8 & 1.94 \\
        \rowcolor{gray!10} \multicolumn{16}{c}{\textsc{Decoding Ablation}} \\
        (10) & FLAN-T5-XL~\cite{chung2024scaling} &         DoRA~\cite{liu2024dora} & 32 & 64 & beam (5) & SAT\textsuperscript{+}\cite{jang2025deep} & Our vis. backbone (20s) & \cmark & Sliding Perceiver & 7.9 & 49.4 & 27.4 & 95.2 & 25.0 & 2.05 \\
        \bottomrule
    \end{tabular}
    }
    \caption{\textbf{Ablation study on all components.} We systematically evaluate the contribution of each model component to SLT performance on the \textsc{BOBSL-Sent} validation set. Starting from the baseline B4 score of 3.4, the proposed modifications progressively improve performance to 7.9, representing doubled performance and highlighting the substantial impact of each component on overall translation quality.}
    \label{tab:app:all_components}
\end{table*}

\subsection{Ablation on all components}
\cref{tab:app:all_components} presents a comprehensive ablation study over all design components that contribute to the observed performance gains. 
All experiments are evaluated on the BOBSL-Sent validation set to quantify the incremental effect of each component.

\newpara{Baseline.}
As a reference, we adopt the LiTFiC model trained on the \textsc{BOBSL-Sent} training split for the SLT task for 10 epochs (Experiment (1) in \cref{tab:app:all_components}). 
The baseline setup is as follows: the language model is LLaMA-8B~\cite{touvron2023llama}, fine-tuned via LoRA~\cite{hu2022lora} (rank $r=4$, $\alpha=16$); decoding is performed with greedy search. 
Training uses sentence-level clips obtained by segmenting the video according to pseudo-subtitle alignments derived from the SAT~\cite{Bull21} model, i.e. each training example is the video region corresponding to one subtitle sentence. 
The visual backbone is pretrained for ISLR on \textsc{BOBSL-SIGN} using an 8K pseudo-gloss vocabulary (3.5M pseudo-gloss samples following~\cite{raude2024}), and the visual backbone is frozen. 
The mapping network is a simple MLP composed of fully connected layers with GELU activations. 

Starting from this baseline, we replace or augment individual components (LLM, decoding, alignment, visual backbone, mapping network) to isolate their contributions. 

\newpara{LLM component ablation.}
Experiment~(2) replaces the LLaMA-8B with FLAN-T5-XL (3B). 
Despite having fewer parameters, FLAN-T5-XL yields consistent improvements across all metrics, demonstrating that a smaller yet instruction-tuned model can serve as a more effective decoder. 
Experiment~(3) applies DoRA~\cite{liu2024dora} finetuning in place of LoRA~\cite{hu2022lora} on top of the FLAN-T5-XL setup, providing additional gains (e.g.\ +0.2 B4), indicating that parameter decomposition further stabilises LLM adaptation. 
In Experiment~(4), we increase the DoRA rank and scaling parameters to 32 and 64, respectively, which leads to a further improvement of +0.7 B4, showing that higher rank adaptations strengthen the generation capability of the model.

\newpara{Alignment ablation.}
In Experiment~(5), we replace the pseudo-subtitles extracted using SAT~\cite{Bull21} with those generated by the stronger SAT\textsuperscript{+}~\cite{jang2025deep} model (53.4 vs.\ 63.8 F1@0.50 on the \textsc{BOBSL-Sent} test set). 
As shown in Experiment~(5), adopting a more accurate alignment model leads to a clear improvement in SLT performance. 
This underscores the importance of high-quality SSA supervision and its direct impact on downstream translation.

\newpara{Feature ablation.}
Experiment~(6) replaces the features obtained from Video-Swin-ISLR~\cite{raude2024} with those extracted by our visual backbone, which takes pose and lip features as input. This substitution yields a 0.5 improvement in the B4 score. Additionally, when training with 20-second video clips (Experiment~(7)) so that the model can access broader visual context, the B4 score improves by a further 0.4.

\newpara{Mapping network ablation.}
Replacing the MLP with the proposed Sliding Perceiver leads to a significant improvement in the B4 score, as shown in Experiment~(8), yielding a gain of +0.9. This demonstrates that the Sliding Perceiver maps video embeddings to text embeddings more effectively.

\newpara{End-to-end training.}
Owing to the lightweight design of the proposed visual backbone, it can be updated jointly with the LLM. As shown in Experiment~(9), finetuning the visual backbone yields a substantial B4 score improvement of 0.8, demonstrating the importance of adapting visual representations for downstream tasks.

\newpara{Decoding ablation.}
Finally, we apply beam search decoding (beam size = 5), as shown in Experiment~(10), which results in a 0.5 improvement in B4 score. Qualitative analysis indicates that the model produces fewer repeated outputs for words in which it has high confidence, reducing redundant predictions.

\section{Qualitative results}
\subsection{Sign language translation}
\begin{figure*}[t]
    \centering
    \begin{subfigure}{0.99\linewidth}
        \centering
        \includegraphics[width=\linewidth]{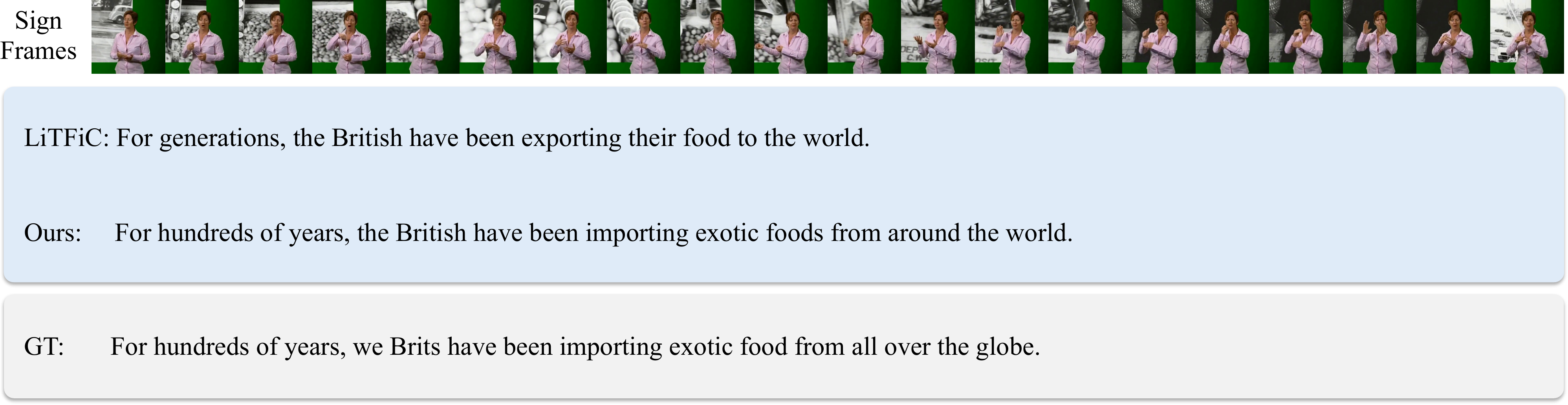}
        \caption{}
        \label{fig:app:translation1}
    \end{subfigure}

    \vspace{1.0em}

    \begin{subfigure}{0.99\linewidth}
        \centering
        \includegraphics[width=\linewidth]{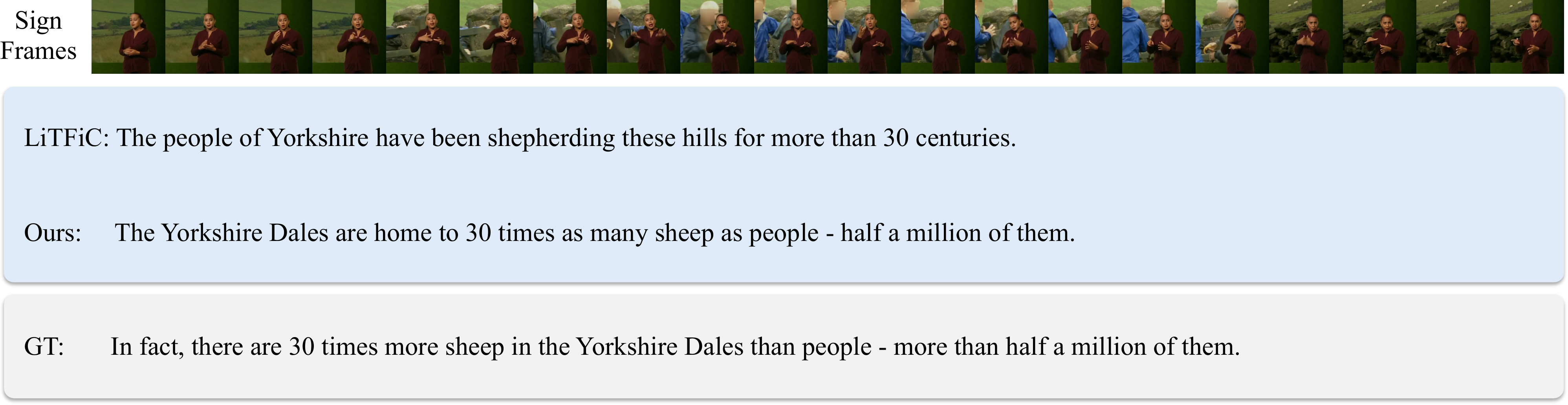}
        \caption{}
        \label{fig:app:translation2}
    \end{subfigure}
    
    \vspace{1.0em}

    \begin{subfigure}{0.99\linewidth}
        \centering
        \includegraphics[width=\linewidth]{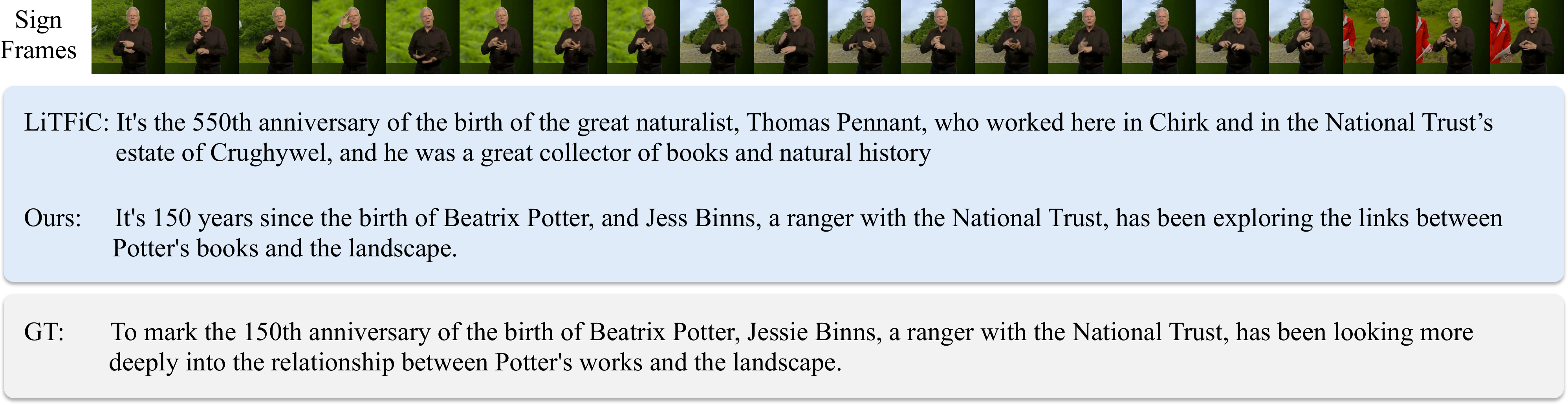}
        \caption{}
        \label{fig:app:translation3}
    \end{subfigure}

    \vspace{1.0em}

    \begin{subfigure}{0.99\linewidth}
        \centering
        \includegraphics[width=\linewidth]{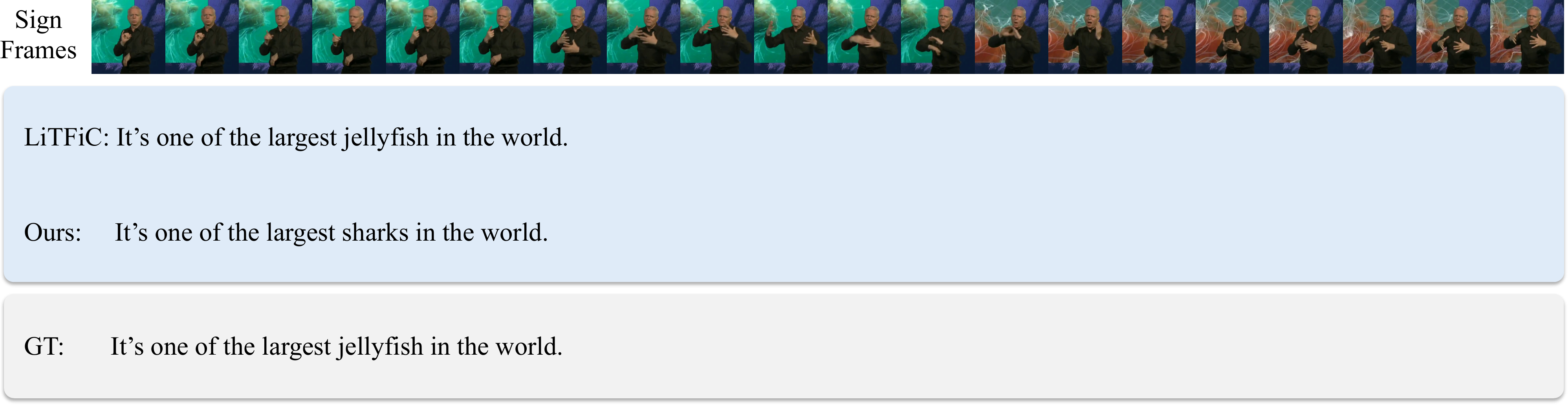}
        \caption{}
        \label{fig:app:translation4}
    \end{subfigure}

    \caption{\textbf{SLT qualitative results on the \textsc{BOBSL-Sent} test set.}}
    \label{fig:app:translation}
\end{figure*}

\newpara{Samples on the BOBSL dataset.}
In \cref{fig:app:translation}, we qualitatively compare the SLT performance of our model with the previous state-of-the-art LiTFiC~\cite{jang2025lost} model. In \cref{fig:app:translation1}, LiTFiC incorrectly predicts \textit{``importing''} as \textit{``exporting.''} In \cref{fig:app:translation2}, while LiTFiC correctly predicts certain keywords such as \textit{``Yorkshire''} and the number \textit{``30,''} the overall interpretation differs from the ground truth. In contrast, our model conveys a similar meaning more accurately. \cref{fig:app:translation3} shows an example of prediction for a long sentence. When the sentence is long, LiTFiC mispredicts the number \textit{``150''} as \textit{``550''} and fails to predict a person’s name correctly, leading to a significantly different meaning. In contrast, our model translates the sentence with a similar meaning while accurately predicting numbers and names.
Finally, \cref{fig:app:translation4} shows a failure case of our model. LiTFiC, which leverages pseudo-glosses, the previous sentence, and background description, correctly predicts the keyword \textit{``jellyfish''} appearing in the background of the video, whereas our model mispredicts it as \textit{``sharks.''} This highlights that in sign language, context beyond the sign movements is crucial and should be appropriately incorporated.

\newpara{Samples on the How2Sign dataset.}
\cref{tab:app:h2s_zero-shot} presents qualitative SLT results on the How2Sign~\cite{duarte2021how2sign} test set. Ours (zero-shot) refers to the model that performs translation using only the large-scale pretrained weights, without any additional training on How2Sign. Ours (finetuned) denotes the model further finetuned on the How2Sign from the large-scale pretrained weights. 
Finally, the Human baseline corresponds to the translations produced by five human signers reported in~\cite{tanzer2024reconsidering}, where each of the five samples comes from a different signer.

\newpara{Samples on the FLEURS-ASL dataset.}
In \cref{tab:app:fleurs}, we provide additional qualitative examples on the FLEURS-ASL zero-shot split with the three zero-shot samples released in the supp.\ mat. of~\cite{tanzer2024fleurs}. We compare the predictions of Gemini 1.5 Pro, FLEURS-SLT, Ours, and the Human baseline.

\begin{table*}
	% \setlength{\tabcolsep}{6pt}
	% \centering
	\resizebox{1\linewidth}{!}
	{
		\begin{tabular}{llp{0.7\linewidth}}
			\toprule
			\multirow[t]{4}{*}{\textbf{1}} & Reference: & Hi, I’m Captain Joe Bruni, and what I want to talk about is how to visually identify prescription drugs. \\
			& Ours (zero-shot): & I'm Capt. Joe Bruni and I'm going to talk about how to visually identify prescription drugs. \\
			& Ours (finetuned): & Now, I'm Captain Joe Bruni, and today we're going to talk about how to visually identify a prescription drug. \\
			& Human \textit{(Signer A)}~\cite{tanzer2024reconsidering}: & I’m Captain Ernie, I want to discuss how to visually identify an Rx drug. \\
			\midrule
			\multirow[t]{4}{*}{\textbf{2}} & Reference: & In boxing you always want to be trying to be moving forward, you want to be trying to be pushed to fight, always trying to be moving forward. \\
			& Ours (zero-shot): & In boxing, you always want to try to move forward. You want to try to punch and fight. Try to move forward. \\
			& Ours (finetuned): & In boxing you always want to try to move forward, you want to try to push fight, you want to try to move forward. \\
			& Human \textit{(Signer B)}~\cite{tanzer2024reconsidering}: & Boxers always want to try to move closer, you want to try to push the fight, try to move closer. \\
			\midrule
			\multirow[t]{4}{*}{\textbf{3}} & Reference: & But one other tip when choosing eye shadow color is actually take a look at color of there eyes. \\
			& Ours (zero-shot): & Another tip when choosing a color for your hair is to look at the color of your eyes. \\
			& Ours (finetuned): & Another tip when picking out the color of the eyebrows is actually to look at the color of the eyes. \\
			& Human \textit{(Signer C)}~\cite{tanzer2024reconsidering}: & A tip when picking the color of your eyeshadow - really look at the color of your eyes. \\
			\midrule
			\multirow[t]{4}{*}{\textbf{4}} & Reference: & You need to consider whether the horse has an illness or an injury. \\
			& Ours (zero-shot): & You want to look at the age of your horse. \\
			& Ours (finetuned): & You need to consider whether your horse has a disease or injury. \\
			& Human \textit{(Signer D)}~\cite{tanzer2024reconsidering}: & Consider if either of your horses have illness or injury. \\
			\midrule
			\multirow[t]{4}{*}{\textbf{5}} & Reference: & If you are looking to buy hosiery for open toe shoes, be it if they are peep toe shoes or if you are looking to wear hosiery with a sandal in the wintertime your best options are to go with hosiery that doesn't have any hem lines or any type of reinforcements. \\
			& Ours (zero-shot): & If it's really hot, it doesn't matter if you're wearing a pair of flip-flops or a pair of sandals in the wintertime, you still need to pick up some heat. \\
			& Ours (finetuned): & So if you want to, it doesn't matter if it's a pep toe shoe or a sandal or a winter time shoe, you still want to pick a heel that has no line, no stitching, no reinforcement. \\
			& Human \textit{(Signer E)}~\cite{tanzer2024reconsidering}: & If you want peep toe shoes or sandals in winter, you should still pick hoir with no lines or reinforcement \\
			\bottomrule
		\end{tabular}
	}
	\caption{\textbf{SLT qualitative results on the How2Sign test set.}
    }
    \vspace{50mm}
	\label{tab:app:h2s_zero-shot}
\end{table*}
\begin{table*}
	% \setlength{\tabcolsep}{6pt}
	% \centering
	\resizebox{1\linewidth}{!}
	{
		\begin{tabular}{llp{0.7\linewidth}}
			\toprule
			\multirow[t]{5}{*}{\textbf{1}} & Reference: & During the 1980s he worked on shows such as Taxi, Cheers, and The Tracy Ullman Show. \\
			& Gemini 1.5 Pro: & I am learning sign language. \\
			& FLEURS-SLT~\cite{tanzer2024fleurs}: & In the 1980s, she worked in theaters like taxesi, cheesy, and tracy. \\
			& Ours: & In the 1980s, he worked on shows like Taxi, Cher, and Tracy Ullman Show. \\
			& Human: & During the 1980s, they worked in shows like Taxi, Cheers, and The Tracy Ullman show. \\
			\midrule
			\multirow[t]{5}{*}{\textbf{2}} & Reference: & The rise of new technologies allows us to see and investigate brain structures and processes never seen before. \\
			& Gemini 1.5 Pro: & You must think! Use your brain! \\
			& FLEURS-SLT~\cite{tanzer2024fleurs}: & There is a new technique to detect brains and vision. \\
			& Ours: & Increasingly, new technology is allowing us to look at the structure of the brain in ways we have never seen before. \\
			& Human: & The rise of new technology allows for investigation of brain structure and processes never seen before. \\
			\midrule
			\multirow[t]{5}{*}{\textbf{3}} & Reference: & The Articles required unanimous consent from all the states before they could be amended and states took the central government so lightly that their representatives were often absent. \\
			& Gemini 1.5 Pro: & It is September. It’s time to go back to school! \\
			& FLEURS-SLT~\cite{tanzer2024fleurs}: & The law requires all states to agree on a standard and that it is a legal requirement. \\
			& Ours: & It requires all states to agree to the same standards, so it can't be amended. \\
			& Human: & The article mandated all states to create a uniform agreement as an amendment to the law for revisions. The states look at the process as minor and many of the state representatives did not make an effort to attend those meetings. \\
			\bottomrule
		\end{tabular}
	}
	\caption{\textbf{SLT qualitative results on the FLEURS-ASL zero-shot split.}
    }
    \vspace{100mm}
	\label{tab:app:fleurs}
\end{table*}

\subsection{Sign-subtitle alignment}
\begin{figure*}[t]
    \centering

    \begin{subfigure}{0.99\linewidth}
        \centering
        \includegraphics[width=\linewidth]{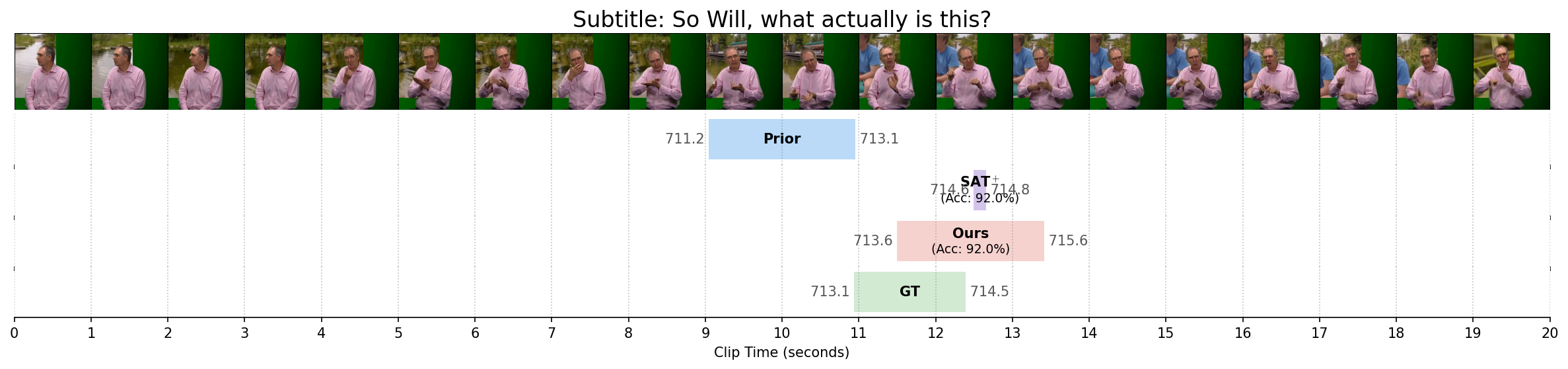}
        \caption{}
        \label{fig:app:alignment_acc1}
    \end{subfigure}

    \vspace{2.0em}

    \begin{subfigure}{0.99\linewidth}
        \centering
        \includegraphics[width=\linewidth]{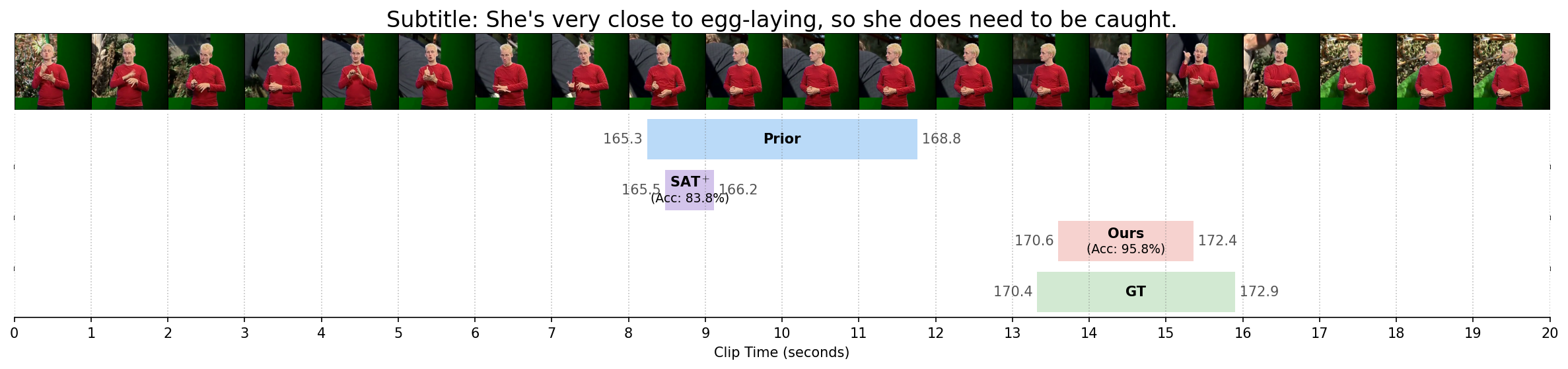}
        \caption{}
        \label{fig:app:alignment_acc2}
    \end{subfigure}
    
    \vspace{2.0em}

    \begin{subfigure}{0.99\linewidth}
        \centering
        \includegraphics[width=\linewidth]{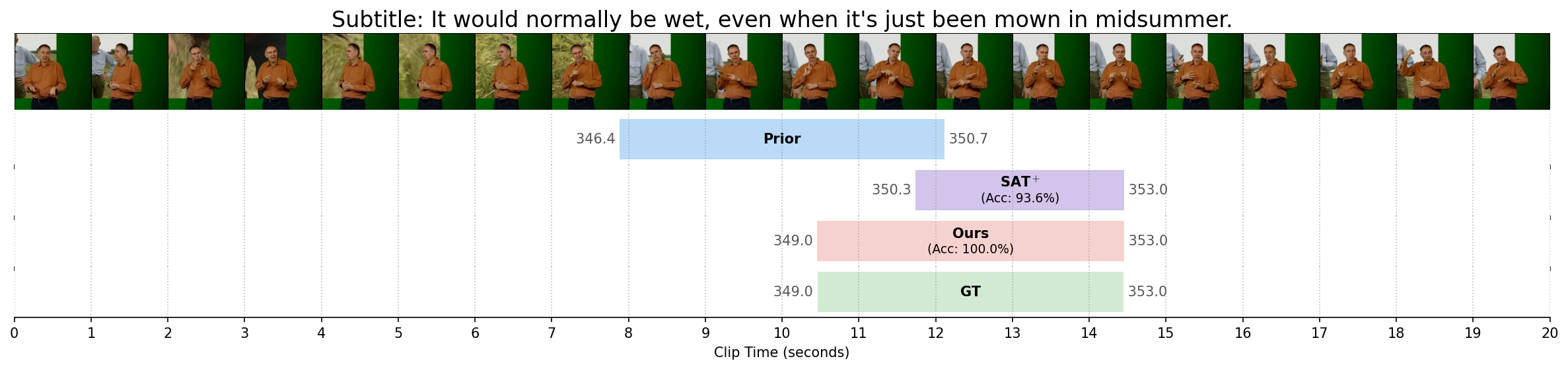}
        \caption{}
        \label{fig:app:alignment_acc3}
    \end{subfigure}

    \vspace{2.0em}

    \begin{subfigure}{0.99\linewidth}
        \centering
        \includegraphics[width=\linewidth]{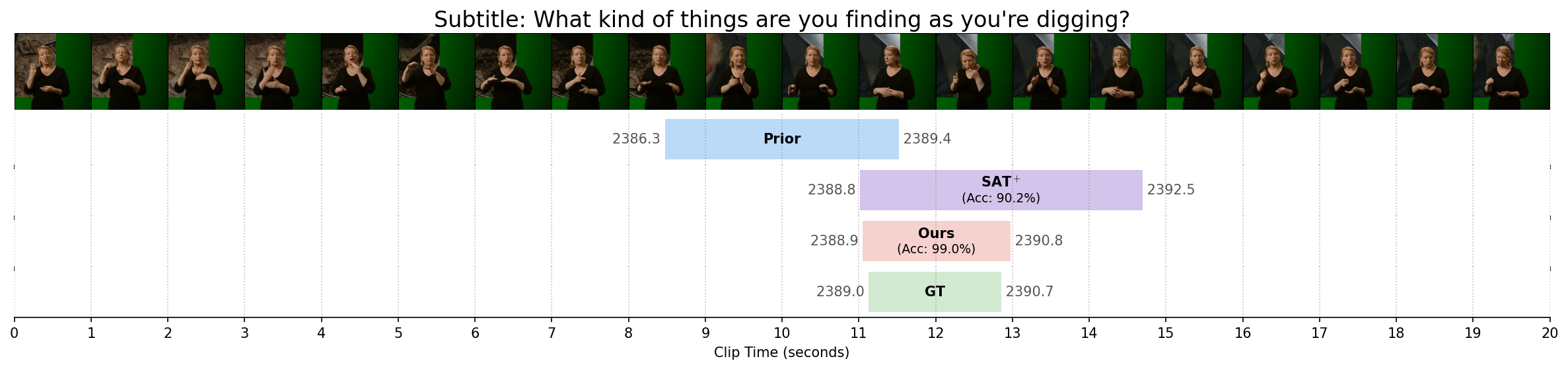}
        \caption{}
        \label{fig:app:alignment_acc4}
    \end{subfigure}

    \caption{\textbf{SSA qualitative results on the \textsc{BOBSL-Sent} test set.}}
    \vspace{10mm}
    \label{fig:app:alignment_acc}
\end{figure*}

We qualitatively compare the alignment performance of our model with the SAT\textsuperscript{+} model in \cref{fig:app:alignment_acc}. In \cref{fig:app:alignment_acc1}, although both models obtain the same frame-level accuracy, SAT\textsuperscript{+} shows no overlap with the ground-truth span, whereas our model achieves around 40\% overlap. This discrepancy arises because frame-level accuracy counts frames outside the ground-truth span as background and incorporates the accuracy on those background frames into the overall score, inflating or masking true alignment performance. Consistent with the results in \appendixref{Sec.~4.4}{\cref{subsec:sota}}, our model attains frame-level accuracy comparable to SAT\textsuperscript{+} but shows substantial improvements in F1 scores, supporting the claim that F1 score is a more reliable metric for alignment performance than frame-level accuracy.
Furthermore, \cref{fig:app:alignment_acc2}, \cref{fig:app:alignment_acc3}, and \cref{fig:app:alignment_acc4} illustrate additional cases where our model produces more accurate alignment predictions than SAT\textsuperscript{+}.

\end{document}